\documentclass[times,twocolumn,review]{elsarticle}

\usepackage{medima}
\usepackage{framed,multirow}

\usepackage{amssymb}
\usepackage{latexsym}

\usepackage{url}
\usepackage{xcolor}

\usepackage{hyperref}
\usepackage{amsmath,amssymb,amsfonts}
\usepackage{algorithm}

\usepackage{algpseudocode}
\usepackage{graphicx}
\usepackage{textcomp}

\usepackage{chngcntr}
\usepackage{placeins}
\usepackage{amsfonts}       
\usepackage{nicefrac} 
\usepackage{amsmath,amssymb} 
\usepackage{amstext}
\usepackage{multirow}
\usepackage{tabu}
\usepackage{svg}
\usepackage{bbm}
\usepackage{diagbox}
\usepackage[english]{babel}
\usepackage{comment}
\usepackage{array}
\usepackage{amstext}
\usepackage{makecell}
\usepackage{caption}
\usepackage{tablefootnote}
\usepackage{babel}
\usepackage{booktabs} 
\usepackage{float}
\usepackage{graphicx}

\newcommand{\revised}[1]{{\color{black} #1}}
\newcommand{\red}[1]{{\color{red} #1}}

\newcommand{\node}{\textbf{g}}
\newcommand{\allnode}{\textbf{G}}
\newcommand{\point}{\textbf{p}}
\newcommand{\pointcloud}{\textbf{P}}
\journal{Medical Image Analysis}

\begin{document}

\verso{Kangxian Xie \textit{et~al.}}

\begin{frontmatter}

\title{Efficient Anatomical Labeling of Pulmonary Tree Structures via Deep Point-Graph Representation-based Implicit Fields}

\author[1,2]{Kangxian \snm{Xie}}
\fntext[fn1]{This work was conducted during K. Xie's research internship at EPFL and Boston College.}
\author[1]{Jiancheng \snm{Yang} \corref{cor1}}

\author[2]{Donglai \snm{Wei}}
\author[3]{Ziqiao \snm{Weng}}
\author[1]{Pascal \snm{Fua}} 
\cortext[cor1]{Corresponding author: 
J. Yang (jiancheng.yang@epfl.ch)}

\address[1]{Computer Vision Laboratory, Swiss Federal Institute of Technology Lausanne (EPFL), Lausanne 1015, Switzerland}
\address[2]{Boston College, Chestnut Hill, MA 02467, USA}
\address[3]{University of Sydney, Camperdown NSW 2050, Australia}


\begin{abstract}
Pulmonary diseases rank prominently among the principal causes of death worldwide. Curing them will require, among other things, a better understanding of the complex 3D tree-shaped structures within the pulmonary system, such as airways, arteries, and veins. Traditional approaches using high-resolution image stacks and standard CNNs on dense voxel grids face challenges in computational efficiency, limited resolution, local context, and inadequate preservation of shape topology. Our method addresses these issues by shifting from dense voxel to sparse point representation, offering better memory efficiency and global context utilization. However, the inherent sparsity in point representation can lead to a loss of crucial connectivity in tree-shaped structures. To mitigate this, we introduce graph learning on skeletonized structures, incorporating differentiable feature fusion for improved topology and long-distance context capture. Furthermore, we employ an implicit function for efficient conversion of sparse representations into dense reconstructions end-to-end. The proposed method not only delivers state-of-the-art performance in labeling accuracy, both overall and at key locations, but also enables efficient inference and the generation of closed surface shapes. Addressing data scarcity in this field, we have also curated a comprehensive dataset to validate our approach. Data and code are available at \url{https://github.com/M3DV/pulmonary-tree-labeling}.

\end{abstract}


\end{frontmatter}


\section{Introduction}
\label{sec:introduction}

In recent years, since pulmonary diseases~\citep{Decramer2008COPDAA,Marcus2000LungCM,Nunes2017AsthmaCA} have become the leading causes of global mortality~\citep{Quaderi2018TheUG}, pulmonary research has gained increasing attention.
In studies related to pulmonary disease, understanding pulmonary anatomies through medical imaging is important due to the known association between pulmonary disease and inferred metrics from lung CT images~\citep{Charbonnier2019AirwayWT,Kirby2019ComputedTT,Qin2019AirwayNetAV,Shaw2002TheRO,doi:10.1148/radiology.219.2.r01ma26498,Shen2014CTbasePA}. 

\begin{figure}
    \centering
    \includegraphics[width=\linewidth]{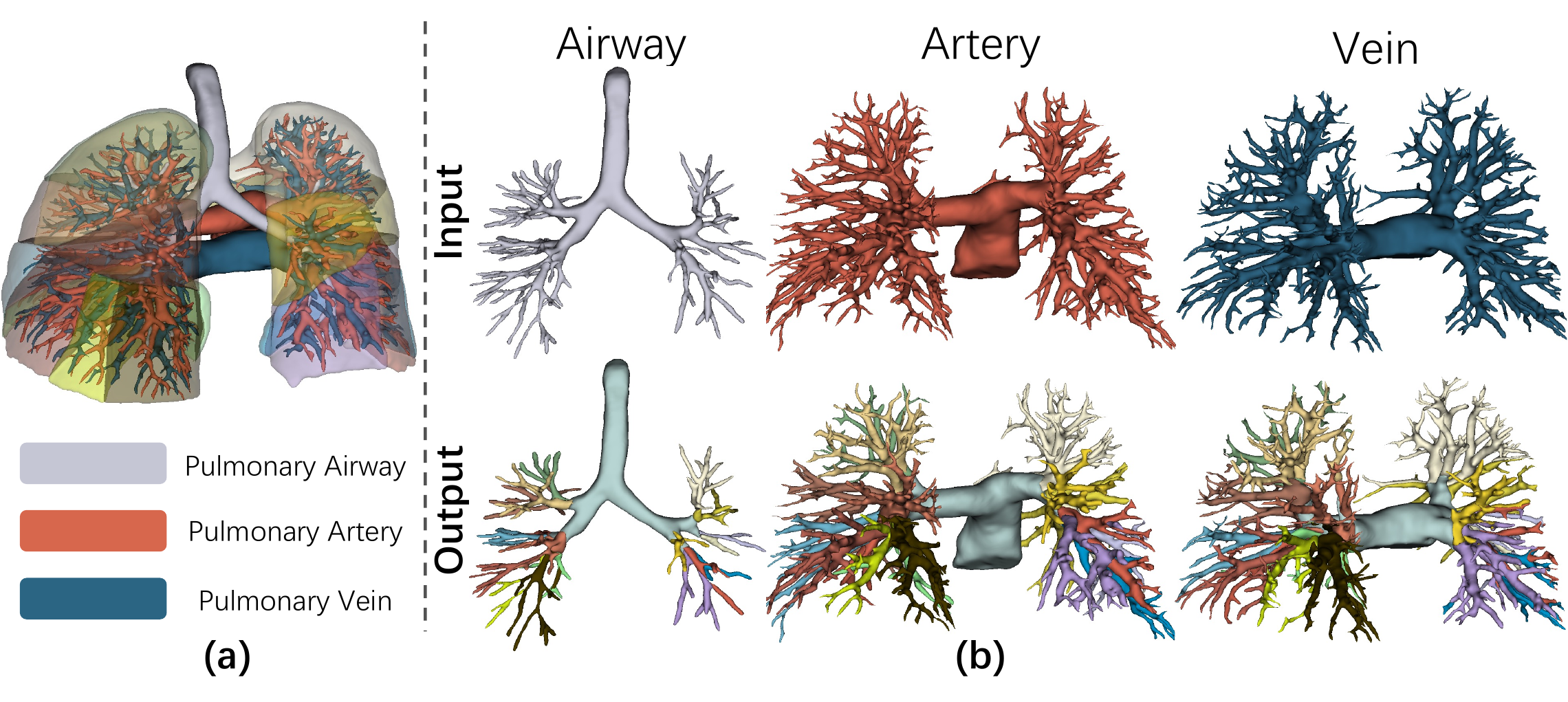}
    \caption{\textbf{Pulmonary Tree Labeling.} (a) The pulmonary tree consists of three anatomic structures (airway, artery and vein). (b) Given a binary volume representing a tree structure as input, we label each voxel into one of 19 classes (18 pulmonary segments + 1 background out of lung) based on branching regions.} 
    \label{fig:dataset}
\end{figure}

The complex tree-shaped pulmonary structures---airways, arteries, and veins, as depicted by Fig. \ref{fig:dataset}---have high branching factors and play a crucial role in the respiratory system in the lung. Multi-class semantic segmentation of the pulmonary trees in volumetric images, where each class represents a specific division or branch of the tree according to the medical definition of the pulmonary segments, is an effective approach to modeling their intricacies, with the challenge primarily lying in the bifurcation locations and distal branches~\citep{TNN}.
In pulmonary tree labeling, the derived quantitative characteristics~\citep{Kirby2019ComputedTT,Smith987,Shaw2002TheRO} not only associate with lung diseases and pulmonary-related medical applications~\citep{doi:10.1148/radiology.219.2.r01ma26498,Shen2014CTbasePA} but are also crucial for surgical navigation~\citep{Qin2019AirwayNetAV}. This work focuses on investigating and developing deep-learning methods for efficient and accurate anatomical labeling of pulmonary airways, arteries, and veins.

\begin{figure*}
    \centering
    \includegraphics[width=0.8\linewidth]{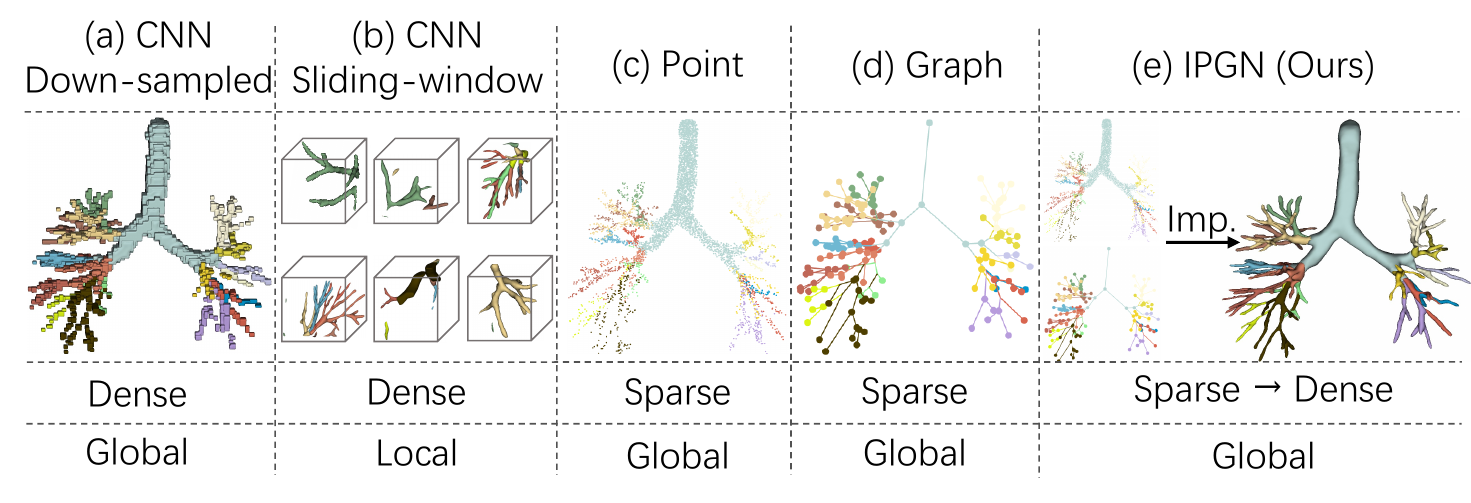}
	\caption{\textbf{A Comparison of Data Representation for Pulmonary Tree Labeling.} The CNN-based voxel methods are either low-resolution (down-sampled) or local (sliding-window). The standard sparse representation like point and graph is global but it is not trivial to reconstruct high-quality dense volume. Our method that combines point, graph, and implicit fields produces high-quality dense reconstruction efficiently.} 
    \label{fig:challenges}
\end{figure*}

Among deep learning approaches, convolutional neural networks (CNN) have become the {\it de facto} standard approach to semantic segmentation~\citep{Zhou2018UNetAN,Zhou2019UNetRS}. One of their strengths is that they yield volumes with well-defined surfaces. However, they are computationally demanding when processing large 3D volumes and often deliver unsatisfactory results when operating at a reduced resolution (Fig.~\ref{fig:challenges}~(a)) or on local patches (Fig.~\ref{fig:challenges}~(b)), either leading to a lack of
details or global context. In contrast, point-cloud is a sparse representation of 3D shapes without usable surface, but it comes with lower computational requirements while preserving global structures (Fig.~\ref{fig:challenges}~(c)). Besides, due to the inherent tree-shaped targets, graph modeling (Fig.~\ref{fig:challenges}~(d)) that preserves the topology and emphasizes learning on key locations is also visible~\citep{TNN,tsinghua,NetherlandXie2022StructureAP}.Nevertheless, extracting usable surfaces from point clouds or graphs is non-trivial.

Considering the challenges mentioned above, as illustrated in Fig.~\ref{fig:challenges}~(e), we would like to develop an optimal automated solution to this task, which is computationally efficient, and reconstruct the 3D model with continuous surface definition and topology preservation, specifically focusing on the key structural points. To this end, we introduce a novel approach that enables deep infusion of skeleton-graph context into the point-based representations, with an implicit method to yield a feature field, which allows for efficient and denoised dense volume reconstruction.
The proposed \emph{Implicit Point-Graph Network~(IPGN)} includes point and graph encoders, combined with multiple \emph{Point-Graph Fusion} layers as \emph{Point-Graph Network~(PGN)} for deep feature fusion and an \emph{Implicit Point Module} to model the implicit surface in 3D, allowing for fast classification inference at arbitrary locations. Thanks to the flexibility of implicit representations, the IPGN trained for pulmonary tree labeling can be further extended to pulmonary segment reconstruction by simply modifying the inference method (Sec.~\ref{sec:extended-app}).

As illustrated by Fig.~\ref{fig:qualitative}, our approach produces high-quality dense reconstructions of pulmonary structures at an acceptable computational cost. To evaluate it quantitatively, we compiled the Pulmonary Tree Labeling (PTL) dataset illustrated in Fig. \ref{fig:dataset}. It contains manual annotations for 19 different components of pulmonary airways, arteries, and vein, which will be made publicly available as a multi-class semantic segmentation benchmark for pulmonary trees. Our method achieves state-of-the-art performance on this dataset while being the most computationally efficient. Our contributions can be summarized as:

\begin{itemize}
    \item We present a new method for analyzing 3D pulmonary tree structures, transitioning from dense voxel to sparse representation for improved memory efficiency and global context understanding.
    \item To enhance topology preservation and connectivity, we propose differentiable point-graph feature fusion on sparse points and skeletonized tree structures.  
    \item It further incorporates an implicit function to efficiently convert sparse representations into dense reconstructions.
    \item We have compiled a comprehensive Pulmonary Tree Labeling (PTL) dataset, achieving superior labeling accuracy over previous methods at both the voxel (overall performance) and graph (key-location performance) levels, while also demonstrating significantly higher efficiency.
\end{itemize}

\section{Related Works}

\subsection{Pulmonary Anatomy Segmentation}
\label{subsec:pultree_label}

In pulmonary-related studies, image-based understanding of pulmonary anatomies is important as metrics inferred from lung imaging have shown to be related to severity, development and therapeutic outcome of pulmonary diseases~\citep{Charbonnier2019AirwayWT,Kirby2019ComputedTT,Qin2019AirwayNetAV,Shaw2002TheRO,doi:10.1148/radiology.219.2.r01ma26498,Shen2014CTbasePA}.

Previously, CNN-based methods have been tailored for comprehension of different pulmonary anatomies, such as pulmonary lobe~\citep{pulanaseg_Gerard2019PulmonaryLS}, airway~\citep{zhang2023multi}. artery-vein~\citep{pulanaseg_Nardelli2018PulmonaryAC,pulanaseg_Qin2020LearningTC} and fissure~\citep{pulanaseg_Gerard2019FissureNetAD}. Among various pulmonary anatomies, tree-shaped pulmonary structures have drawn a lot of research attention.
For pulmonary airway tree, not only is proper segmentation crucial for surgical navigation~\citep{Qin2019AirwayNetAV}, segmentation-derived attributes like airway counts~\citep{Kirby2019ComputedTT}, thickness of wall~\citep{Smith987} and morphological changes~\citep{Shaw2002TheRO} also have association with lung diseases. For pulmonary vasculature including arteries and veins, quantitative attributes extracted from segmentation are also commonly applied in multiple pulmonary-related medical applications like emboli detection~\citep{doi:10.1148/radiology.219.2.r01ma26498} and hypertension~\citep{Shen2014CTbasePA}.

Previous works specifically on pulmonary tree segmentation either apply graph-only modeling or leverage graph-CNN multi-task learning. A Tree-Neural Network~\citep{TNN} leverages handcrafted features on a tailored hyper-graph for pulmonary bronchial tree segmentation to address the inherent problem of overlapping distribution of features. A recent work~\citep{NetherlandXie2022StructureAP} on airway segmentation proposes graph modeling to incorporate structural context to local CNN feature from each airway branch. Yet, both works merely provide labeling to pulmonary tree branches that are pre-defined at pixel level, and thus are not for semantic segmentation, where defining accurate borders between neighboring branches remains a challenge. Applicable to binary or raw images of pulmonary structures, SG-Net~\citep{tsinghua} employs CNN features for detecting of landmarks, constructing graphs for CNN-graph multi-task learning. 

Although these methods are graph-based, the graph construction procedure varies. While one treats each pre-defined branch as a node~\citep{NetherlandXie2022StructureAP}, disrupting the original spatial structure, another is parameter-based~\citep{TNN}, causing the quality of the constructed tree highly dependent on the parameter selection, and finally, the SG-Net~\citep{tsinghua} established its graph node by learned landmark prediction, whose structural quality can not be ensured. In our setup, the skeleton graphs are based on a thinning algorithm~\citep{LEE1994}, with no modeling or hyper-parameters tuning involved, and all spatial and connection relationships between tree segments are acquired directly from the original dense volume (Fig. \ref{fig:dataset}). Additionally, as CNN methods incur memory footprint that grows in cubical, they are expensive when facing large 3D volumes. 

\subsection{3D Deep Learning}
\label{subsec:geometric}

Deep learning on 3D geometric data is a specialized field that focuses on adapting neural networks to manage non-Euclidean domains, such as point clouds, graphs (or meshes), and implicit surfaces. 3D CNNs for voxel representations can also handle shapes~\cite{wu20153D}, but encounter limitations due to their dense nature, leading to high memory and computational demands. As a result, these representations often necessitate downsampling for modeling at lower resolutions (Fig.~\ref{fig:challenges} (a)), or the use of sliding-window techniques (Fig.~\ref{fig:challenges} (b)), which unfortunately restrict them to capturing only local context, thereby failing to effectively utilize global context.

Point-based methods have emerged as a novel approach for 3D geometric modeling. As sparse 3D data~(Fig.~\ref{fig:challenges} (c)), point cloud can be modeled in a variety of ways, such as multi-layer perception~\citep{Qi2016PointNetDL,pointnet++}, convolution~\citep{Li2018PointCNNCO,Liu2019PointVoxelCF_pvcnn}, graph~\citep{Lin2020ConvolutionIT_pointcloudgraph} and transformer~\citep{yang2019modeling,zhao2021point}. Point-based methods have also been validated in the medical image analysis domain~\citep{Yang_2021_ribsegpointcloud,jin2023ribseg}. However, point cloud, as sparse representation of 3D shape, does not provide a closed surface.

Since the initial introduction~\citep{GCN}, graph learning has become a powerful tool for analyzing graph-structured data~(Fig.~\ref{fig:challenges} (d)), being widely applied to medical images analysis~\citep{TNN,NetherlandXie2022StructureAP,tsinghua} or bioinformatics~\citep{Jumper2021HighlyAP_alphafold}. Meshes represent a specialized form of graph, typically characterized by a collection of triangular faces to depict the object surfaces, and are extensively employed in the field of graphics. Additionally, they also maintain some applications in the domain of medical imaging~\citep{wickramasinghe2020voxel2mesh,wickramasinghe2022weakly}.
While point-based methods enable efficient sparse-representation learning on dense volumetric data, graph learning is lightweight and learns structural context within geometric data. Combining point and graph methods is advantageous in our task. However, extracting surfaces (or dense volumes) from sparse point-based or graph-based prediction is non-trivial, which leads us to introduce implicit surface representation to address this issue~(Fig.~\ref{fig:challenges} (e)).

Deep implicit fields / functions have been successful at modeling 3D shapes~\citep{imp:Chen2018LearningIF,imp:Chibane2020ImplicitFI,imp:Huang2022RepresentationAgnosticSF,imp:Mescheder2018OccupancyNL,imp:Park2019DeepSDFLC,imp:Peng2020ConvolutionalON,yang2022implicitatlas}. Typically, the implicit function predicts occupancy or signed distance at continuous locations, thus capable of reconstructing 3D shapes at arbitrary resolutions~\citep{Kuang2022WhatMF}. Moreover, implicit networks could be trained with randomly sampled points, which lowers the computation burden during training. These characteristics suggest that implicit fields have inherent advantages when reconstructing the sparse prediction of pulmonary tree labeling into dense.


\begin{figure}
    \centering
    \includegraphics[width=\linewidth]{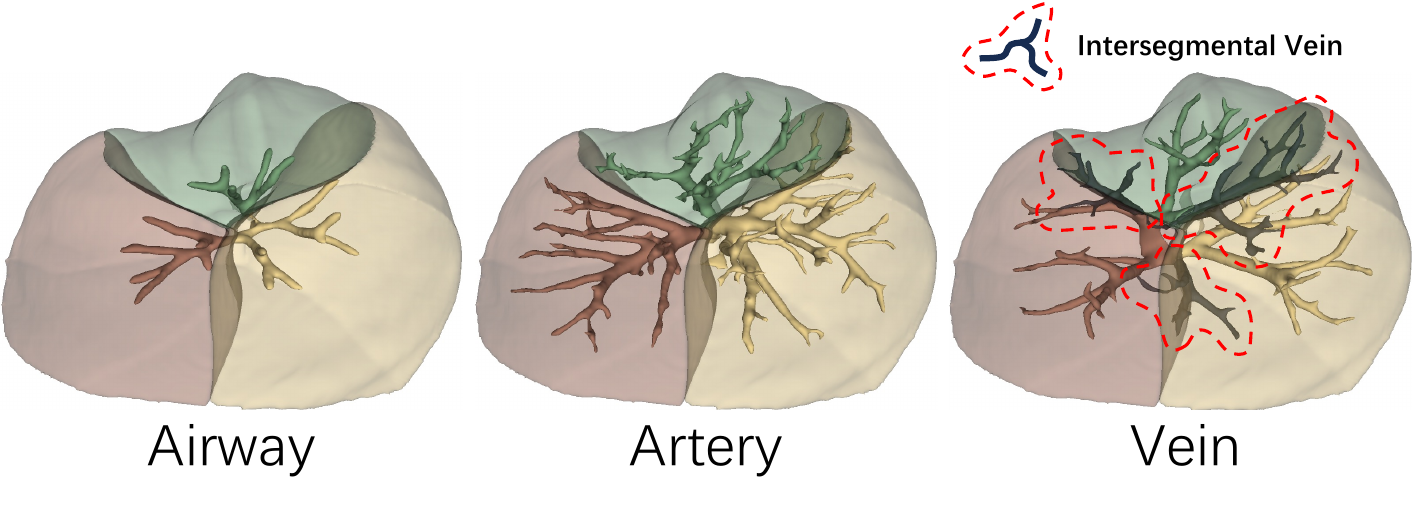}
	\caption{\textbf{Anatomy of Pulmonary Trees and Pulmonary Segments.} Each pulmonary tree branch corresponds to a pulmonary segment. The intersegmental vein is highlighted in red, which lies along the pulmonary segment border.} 
    \label{fig:problem_formulation}
\end{figure}

Additionally, when evaluating the segmentation performance, we consider the presence of intersegmental veins and the potential ambiguity in their class assignment. Intersegmental veins are veins that lie along the border between two neighboring pulmonary segments~\citep{Oizumi2014TechniquesTD_inter_vein}, highlighted in Fig. \ref{fig:problem_formulation}. As pulmonary tree branches are involved in the boundary definition of pulmonary segments~\citep{Kuang2022WhatMF}, intersegmental veins pose an inherent challenge in their class definition. To address this issue, we mask out the intersegmental veins during the evaluation and only focus on segmentation of the pulmonary airway, artery, and veins within each individual segment.
 
\begin{figure}
    \centering
    \includegraphics[width=\linewidth]{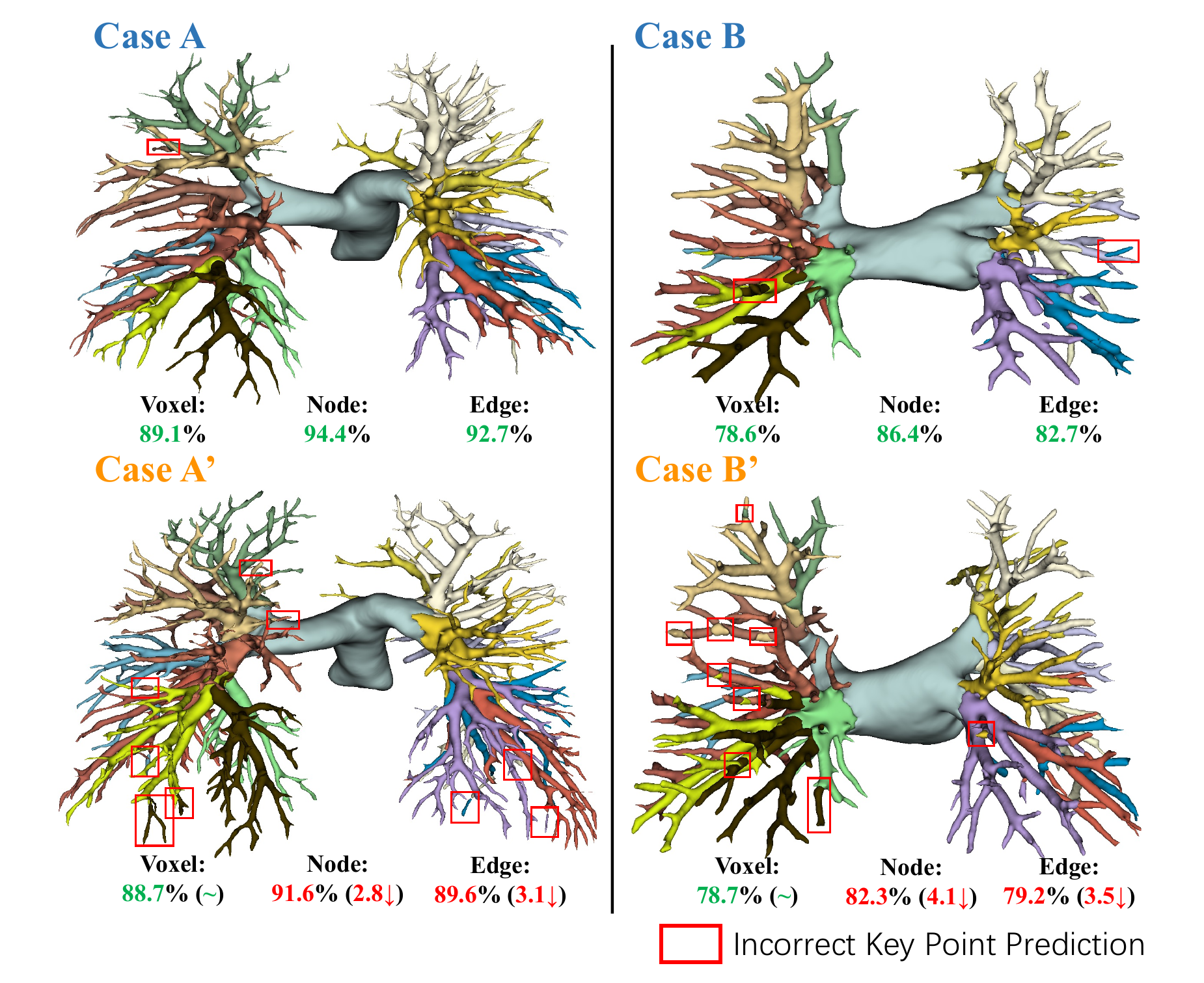}
	\caption{\textbf{Sample Illustration with Voxel-level and Graph-level Metrics.} Due to the large number of voxels, voxel-level metric often overlooks labeling at key regions. Given two test samples with similar voxel performance, the sample with better graph-level performance ({\color[HTML]{2596be} Case A and B}) tends to perform better at key points than its counterpart ({\color[HTML]{e28743} Case A' and B'}). }
\label{fig:metrics}
\end{figure}

\section{Problem Formulation}
\label{sec:problem_formulation}

\subsection{Pulmonary Tree Labeling}
\label{subsec:problem_formulation}
In this study, we address the pulmonary tree labeling problem in 3D CT images. Specifically, given a binary volumetric image of a pulmonary tree, our objective is to provide an accurate 19-class (18 foreground classes + 1 background) segmentation of the pulmonary airway, artery, and vein trees into segmental-level branches and extra-pulmonary components, demonstrated in Fig.~\ref{fig:dataset} and Fig.~\ref{fig:problem_formulation}. Each foreground pixel will be assigned to its respective semantic class. 

While this task reconstructs a high-resolution 3D shape with dense annotations, the primary challenge lies in the correct labeling of key branching points, and distal end points~\citep{TNN}. As the pixels contained in the 3D image have large quantities, learning signals targeting these key locations in the tree structures is overpowered. In Fig.~\ref{fig:metrics}, a few test samples illustrate that overall voxel-level dice performances do not translate to performances at key locations due to the large volume of pixels that could be easily classified. Therefore, another key component of this labeling problem is to ensure a particularly precise annotation at these important locations. To this end, we preprocess the volume and derive an approximate representation of key points as nodes in a skeleton graph (Sec.~\ref{subsec:dataset_preprocess}) and perform graph node/edge classification for evaluation.

\subsection{Dataset}
\subsubsection{Overview}
\label{subsec:dataset_overview}
From multiple medical centers, we compile the Pulmonary Tree Labeling (PTL) dataset containing annotated pulmonary tree structures for 800 subjects. For each subject, there are 3 volumetric images containing its pulmonary airway, artery, and vein, illustrated in Fig. \ref{fig:dataset} (b). In each volume, the annotations consist of 19 classes, where labels 1 to 10 represent branches located in the left lung, labels 11 to 18 represent those in the right lung, and class 19 represents extra-pulmonary structures.

All 3D volumes have shapes $N \times 512 \times 512$, where $512 \times 512$ represents the CT slice dimension, and $N$ denotes the number of CT slices, ranging from 181 to 798. Z-direction spacings of these scans range from 0.5mm to 1.5mm. Manual annotations are produced by a junior radiologist and verified by a senior radiologist. During the modeling process, the images are converted to binary as input, and the original annotated image is the prediction target. The dataset is randomly split into 70\% train, 10\% validation, and 20\% test samples.

\subsubsection{Skeleton Graph}
\label{subsec:dataset_preprocess}

As discussed in Sec.~\ref{subsec:problem_formulation}, other than the voxel-level labeling on the entire image foreground, labeling of key branching points is considered a crucial component of the task. As the branching key points and distal endpoints are easily overwhelmed by image pixels in quantity, we construct a graph for each 3D volume, with nodes as approximate representations of the key points for evaluation~(Fig.~\ref{fig:challenges} (d)).

The skeleton graph dataset is derived from the PTL dataset. Given a 3D volume containing a medical shape, such as a tree-shaped pulmonary structure, we utilize a software application vesselvio~\citep{vesselvio}, which applies a well-known thinning algorithm~\citep{LEE1994}, to derive the shape's centerlines, which are used to construct a connectivity-preserving skeleton tree graph in 3D space, illustrated by Fig. \ref{fig:challenges} (d). In the resulting tree graphs, centerlines become edges that represent anatomical connections, centerline intersections and distal endpoints become nodes that represent key points in the pulmonary structures that require specific attention. \revised{In this manually derived graph dataset, we specialized a labeling procedure to ensure annotation consistency between the software-generated graph and the original volume. In this procedure, we start by labeling edges before node so that the deterministic edge label could improve the quality of node labels that are non-deterministic at class intersections. More specifically, the annotation process can be summarized as the following:} 

\begin{figure}
    \centering
    \includegraphics[width=\linewidth]{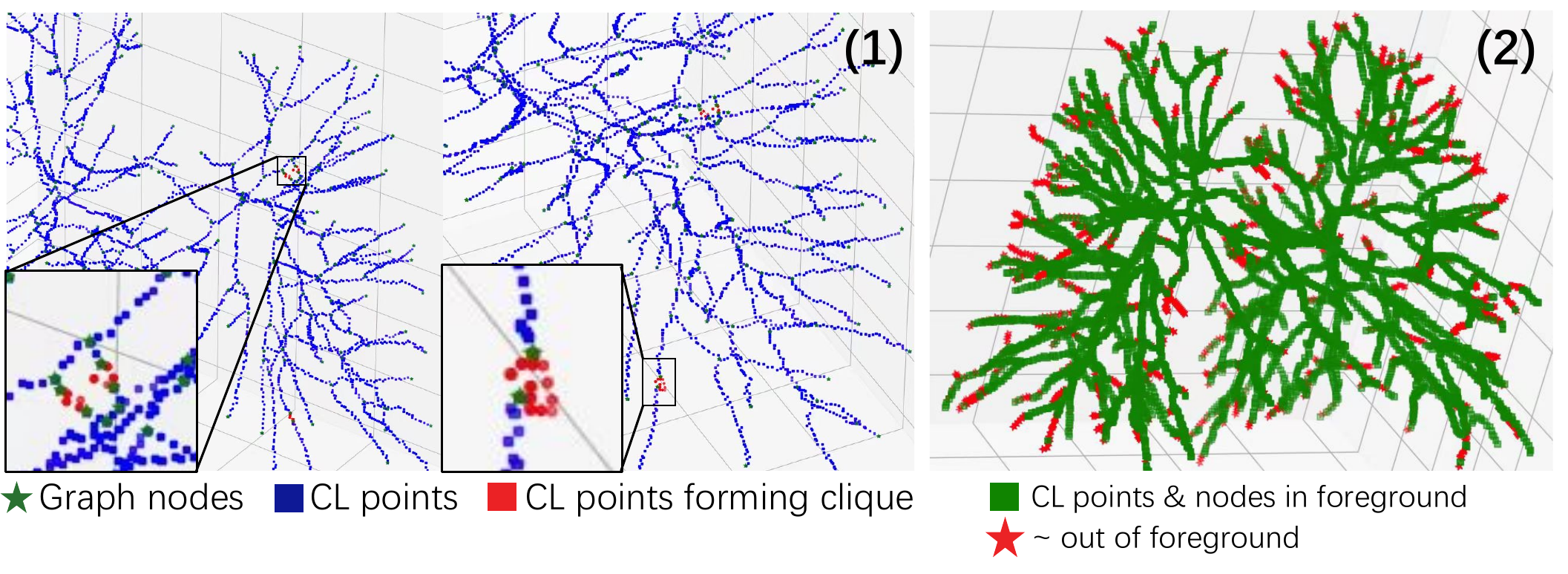}
	\caption{\revised{\textbf{Skeleton Graphs Directly from~\cite{vesselvio} and Two Types of Imperfections.} (1)~The centerline (CL) points, which represent the path of the tree branch as discrete and connected coordinates, might contain cliques. (2)~The coordinates of leaf graph nodes might fall outside the foreground volume.}}
\label{fig:graph_con}
\end{figure}
\revised{

\paragraph{Graph Simplification} From software~\citep{vesselvio} output, for each edge $e$, inter-connected point coordinates sparsely representing the centerline of $e$ are also recorded~(Fig.~\ref{fig:graph_con} CL points) and they are key to the subsequent label assignment. However, due to software inconsistency, cliques occur as displayed in Fig.~\ref{fig:graph_con} (1). To best prepare for the subsequent labeling, we remove the cliques by only keeping the shortest path between two nodes.

\paragraph{Edge Labeling} For any $e \in E$, we recover the labels of the points on the centerline from the original volume. Then, majority voting is applied on the centerline labels to determine the final label for $e$.

\paragraph{Node Labeling} If a node is a leaf node, the node inherits the label of that edge. This has advantage over seeking for label directly from the volume because due to software inconsistency, the coordinates of leaf nodes might not fall into volume foreground, illustrated in Fig.~\ref{fig:graph_con} (2). For nodes with multiple edges, majority voting is applied on the labels of all connected edges to determine the node's label. 

}

\subsection{Evaluation Metrics}
\label{sec:metrics}

This procedure enables a robust ground truth annotation of the graph, reflecting both the local and global structural information contained within the original 3D volume.
Therefore, the task within this graph dataset involves performing a 19-class classification for both graph nodes and edges.

For performance evaluation, micro-averaged dice scores are used as metrics due to class imbalance. For voxel-level results on dense volume, \revised{we employ the dice score to assess the overlap between the predicted segmental-level branch and the ground truth branch}, providing a measure of similarity. As illustrated in Fig.~\ref{fig:metrics}, voxel-level results overlook performances at some critical locations, in which volumes are significantly less. To bridge this discrepancy, we also evaluate the performance at the skeleton graph-level (Sec.~\ref{subsec:dataset_preprocess}), to estimate performance at key points of the tree structure. For graph-level node and edge classification, we also utilize dice as metrics. The dice score can be used to assess the similarity of the predicted graph structure with the ground truth graph structure.

\section{Methodology}
\label{sec:method}

\begin{table*}
\resizebox{\linewidth}{!}{%
\begin{tabular}{c|c}
\hline
\textbf{Notation} & \textbf{Definition} \\ \hline
\multicolumn{2}{c}{\textbf{Point-Graph Feature Fusion}} \\ \hline
$\pointcloud = \{\point_{1},..., \point_{m} \} \in \mathbb{R}^{M\times3}$ & The original point cloud with $M$ points. \\
$\point = \{x,y,z\} \in \mathbb{R}^{3}$ & A single point in the original point cloud $\pointcloud$, $\point \in \pointcloud$. \\
$\point^{(i)} = \{\point_{1}^{(i)},..., \point_{m}^{(i)} \} \in \mathbb{R}^{M\times128}$ & Features of the original point cloud $\textbf{P}$ after the $i$-th layer. \\
$\point^{(i)} \in \mathbb{R}^{128}$ & Feature of a single point $\point$ after the $i$-th layer, $\point^{(i)} \in \pointcloud^{(i)}$. \\ \hline

$\allnode = \{\node_{1},..., \node_{n} \} \in \mathbb{R}^{N\times3}$ & All $N$ nodes in a skeleton graph. \\
$\node \in \mathbb{R}^{3}$ & A single node in the graph node set $\allnode$, $\node \in \allnode$. \\
$\allnode^{(i)} = \{\node_{1}^{(i)},..., \node_{n}^{(i)} \} \in \mathbb{R}^{N\times128}$ & Features of the graph node set $\allnode$ after the $i$-th layer. \\
$\node^{(i)} \in \mathbb{R}^{128}$ & Feature of a single node $\node$ after the $i$-th layer, $\node^{(i)} \in \allnode^{(i)}$. \\ \hline

$\mathcal{F}_{1}: \mathbb{R}^{128} \rightarrow \mathbb{R}^{128}$ & An MLP for ball queried points for graph branch. \\
$\mathcal{F}_{2}: \mathbb{R}^{256} \rightarrow \mathbb{R}^{128}$ & An MLP for graph feature propagation in point branch. \\
$\mathcal{G}: \mathbb{R}^{N \times 128} \rightarrow \mathbb{R}^{N \times 128}$ & A GNN to update graph node feature after $\mathcal{F}_{1}$. \\ \hline

\multicolumn{2}{c}{\textbf{Implicit Dense Volume Reconstruction}} \\ \hline
$\point_{q} = \{x_{q},y_{q},z_{q}\}$ & An arbitrary query point, $\point_{q} \in$ or $\not\in \pointcloud$. \\
$\point_{k} \in \mathbb{R}^{3}$ & $\point_{q}$'s $k$-th nearest neighbor point in the original point cloud $\pointcloud$, $\point_{k} \in \pointcloud$. \\
$\point_{k}^{(i)} \in \mathbb{R}^{128}$ & $\point_{k}$'s feature from $i$-th layer point features $\pointcloud^{(i)}$. \\
$\textbf{z}_{k} = \{\point^{(0)}_{k} \frown \point^{(1)}_{k} \frown ... \frown \point^{(i)}_{k}\} \in \mathbb{R}^{128\times(i+1)}$ & $\point_{k}$'s multi-stage feature by concatenating multi-stage point features. \\
$\mathbf{z}_{q} \in \mathbb{R}^{128\times(i+1)}$ & $\point_{q}$'s multi-stage feature using feature propagation from $\{\textbf{z}_{1}, \textbf{z}_{2}, ..., \textbf{z}_{k}\}$. \\ \hline

$\mathcal{H}: \mathbb{R}^{128\times(i+1)} \rightarrow \mathbb{R}^{19}$ & An MLP to project the multi-stage features $\textbf{z}_{q}$ of a query point $\point_{q}$ to prediction. \\ \hline
\end{tabular}%
}
\caption{\revised{\textbf{Notations and Corresponding Descriptions.} This table presents the notations used to represent point and graph elements within the architecture. The notations are organized into two stages: Point-Graph Feature Fusion and Implicit Dense Volume Reconstruction.}}
\label{tab:notation}
\end{table*}

Our objective is centered around the anatomical labeling of pulmonary trees. That is to say, given an input of binary volumes of a pulmonary tree---derivable from manual annotations or model predictions---we aim to execute a 19-class semantic segmentation on every non-zero voxel. However, diverging from standard methodologies reliant on CNNs, we down-sample the raw dense volume to a point cloud while concurrently extracting a skeleton graph~(Sec.~\ref{subsec:dataset_preprocess}). Our approach (Sec.~\ref{sec:ipgn}) engages in representation learning individually on the sparse representations of both the point cloud and the skeleton graph, and a fusion module is employed to perform deep integration of point-graph representations (Sec.~\ref{sec:pgf}). Ultimately, to reconstruct the predictions based on sparse representations back to dense, we introduce implicit fields to facilitate efficient reconstruction (Sec.~\ref{sec:ipm}). \revised{To improve the readability of methodology, we describe the notations in Table~\ref{tab:notation}.}

\subsection{Implicit Point-Graph Network Architecture} 
\label{sec:ipgn}

\begin{figure*}[tb]
    \centering
    \includegraphics[width=\linewidth]{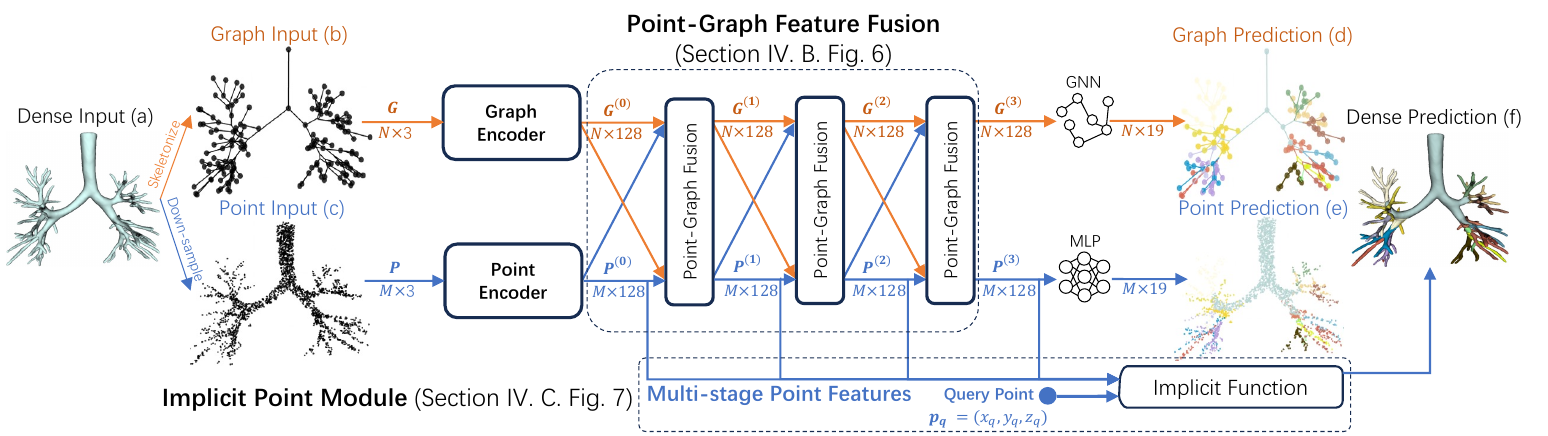}
	\caption{\textbf{Overview of the Proposed Implicit Point-Graph Network (IPGN) for Pulmonary Tree Labeling.} The pipeline pre-processes dense volume to graph and point cloud input. The \emph{Point-Graph Fusion} layers enhance point features with graph context with differentiable feature fusion learning, and the \emph{Implicit Point Module} produces dense prediction based on the deep sparse representation efficiently.}
\label{fig:architecture}
\end{figure*}

Given a binary volumetric image of a pulmonary tree (Fig. \ref{fig:architecture} (a)), a graph (Fig. \ref{fig:architecture} (b)) is constructed with VesselVio~\citep{vesselvio} from the original volume following the procedure in \ref{subsec:dataset_preprocess}, and a set of points ($6k$) are randomly sampled from the foreground voxels to construct a point cloud (Fig. \ref{fig:architecture} (c), \revised{referred to as original point cloud}). While the \revised{original} point cloud is a sparse and downsampled version of the foreground volume, the graph represents a skeleton of the pulmonary tree.

We first introduce the general notation rule for both point and graph elements. While the coordinates of $M$ points and $N$ graph nodes are represented as $\pointcloud$ and \allnode, single point or graph element is expressed as $\point$ and $\node$, where $\pointcloud = \{\point_{1}, \point_{2}, ..., \point_{m}\}$, $\allnode = \{\node_{1}, \node_{2}, ..., \node_{n}\}$. The superscript notation $\point^{(i)}$ represents an element's feature at the $i$-th network layer.

At input, the 3-dimensional $\{x,y,z\}$ point coordinates, $\pointcloud$ $\in$ $\mathbb{R}^{M\times3}$ and graph nodes, $\allnode$ $\in$ $\mathbb{R}^{N\times3}$ are utilized as initial feature.  We use a point neural network, and a graph neural network as initial feature encoders, from which we extract a 128-dimensional intermediate feature for each point and graph node, expressed as $\pointcloud^{(0)}$ $\in$ $\mathbb{R}^{M\times128}$ and $\allnode^{(0)}$ $\in$ $\mathbb{R}^{N\times128}$.

Subsequently, initial features from both branches $\pointcloud^{(0)}$ and $\allnode^{(0)}$ are incorporated within one or multiple \emph{Point-Graph Fusion} layers, which allow for two-way feature integration based on feature propagation~\citep{fp_Rossi2021OnTU} and ball-query\&grouping~\citep{pointnet++}. Let the input to a Point-Graph Fusion layer be defined as $\pointcloud^{(i-1)}$ and $\allnode^{(i-1)}$, the feature out of the fusion layer is $\pointcloud^{(i)}$ and $\allnode^{(i)}$. The last Point-Graph Fusion layer outputs $\pointcloud^{(l)}$ and $\allnode^{(l)}$ after $l$ {Point-Graph Fusion} layers for deep feature fusion. Finally, a lightweight MLP network and a GNN projects the fusion feature to 19-dimensional vectors for graph (Fig. \ref{fig:architecture}~(d)) and point predictions (Fig. \ref{fig:architecture}~(e)).

An \emph{Implicit Point Module} is further introduced to reconstruct the dense volumes, which consists of a feature propagation process and an MLP network. As features are extracted by the Point-Graph Network, the Implicit Point Module leverages the extracted multi-stage point features for fast dense volume segmentation.\revised{We define the points that are not necessarily in the original point cloud, but are in the foreground of the tree volume, thus requiring classification, as query points.} Given a query point $\point_{q}$ with at arbitrary coordinates, the module locates $\point_{q}$'s $k$-nearest point elements from the \revised{original} point cloud: $\{\point_{1}, \point_{2}, ..., \point_{k}\}$, and extracts their multi-stage features $\{\textbf{z}_{1}, \textbf{z}_{2}, ..., \textbf{z}_{k}\}$ from the backbone Network for feature propagation into a multi-stage representation $\textbf{z}_{q}$ of the query point $\point_{q}$. After propagating the point feature $\textbf{z}_{q}$, the MLP network $\mathcal{H}$ is utilized to make class predictions. By applying this process to all foreground points, we can efficiently generate a dense volume reconstruction (Fig.~\ref{fig:architecture} (f)).

To avoid naming ambiguity, we refer to the aforementioned complete network as \emph{Implicit Point-Graph Network (IPGN)}, and that sans the implicit part \revised{as the backbone network,} \emph{Point-Graph Network (PGN)}.

\subsection{Point-Graph Feature Fusion}
\label{sec:pgf}

\begin{figure*}
    \centering
    \includegraphics[width=0.9\linewidth]{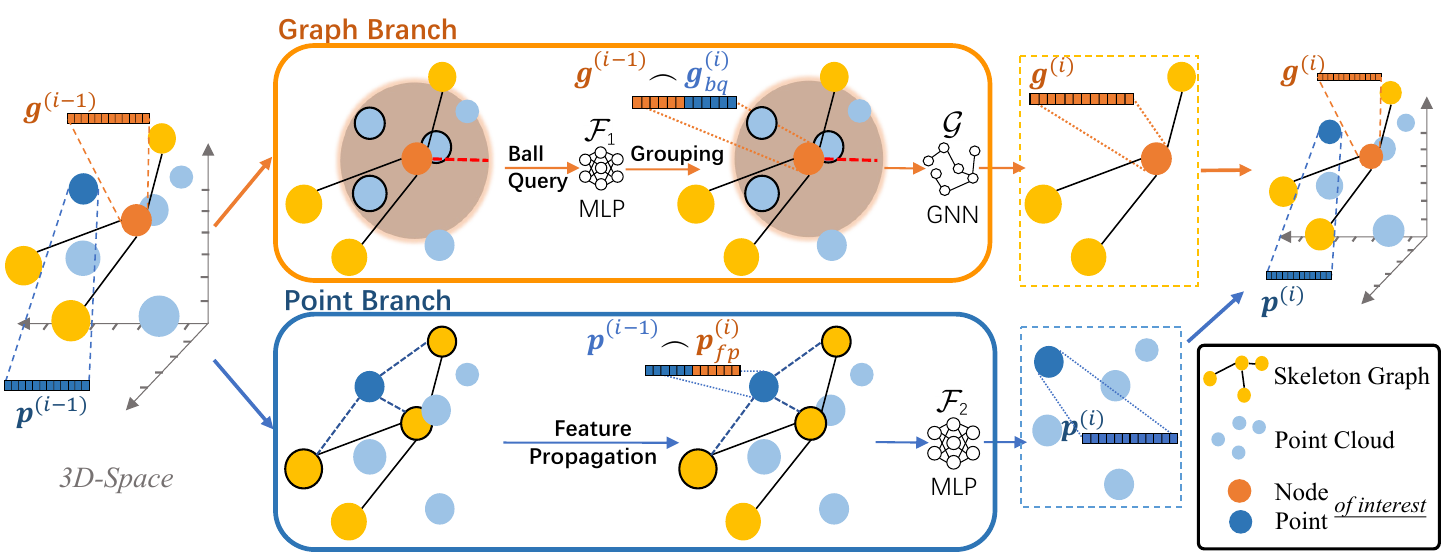}
	\caption{\textbf{The Detailed Operation of a Point-Graph Fusion layer.} The figure only shows the operations for one node, and one point, which are highlighted as Node/Point \emph{of interest}. In reality, this process is parallel for all nodes and points. Here, $\mathcal{F}_{1}$ and $\mathcal{F}_{2}$ are MLP networks while $\mathcal{G}$ represents a graph neural network.}
\label{fig:pg_layer}
\end{figure*}

The essence of our point-graph fusion learning approach lies in leveraging coordinate information as a basis for querying neighboring elements in the opposite branch. To achieve feature integration, We adopt ball-query \& grouping for point-to-graph feature merging and feature propagation for graph-to-point feature merging. 

For the ball-query \& grouping method in the $i$-th Point-Graph Fusion layer, a graph node $\node$ searches for all opposite point elements within a given ball with radius $r$ as $\{\point_{1}, \point_{2}, ..., \point_{b} \}$. Then, an MLP module $\mathcal{F}_{1}:\mathbb{R}^{D} \rightarrow \mathbb{R}^{D}$ independently project the point feature vectors to an updated representation of all queried points,
Then, a feature-wise max-pooling layer aggregates all updated point features as point representation of the node $\node$, expressed as:

\begin{equation}
\node^{(i)}_{bq} = \max_{j} \left(\mathcal{F}_{1}( \point^{(i-1)}_{j}) \right)
\end{equation} 

Subsequently, the ball-queried feature $\node^{(i)}_{bq}$ is combined with the current feature $\node^{(i-1)}$ before using a shallow graph attention network~\citep{GAT} $\mathcal{G}: \mathbb{R}^{2D}\rightarrow \mathbb{R}^{D_{next}}$ to perform graph convolution for feature fusion, resulting an updated feature representation of the node, $\node^{(i)}$ $\in$ $\mathbb{R}^{D_{next}}$, as input to the next Point-Graph Fusion layer.

For feature fusion from graph to point, feature propagation is utilized. In the process, each query point \emph{p} with feature $\point^{(i-1)}$ $\in$ $\mathbb{R}^{D}$ at the $i$-th fusion layer locates its $k$-nearest graph nodes $\{\node_{1}, \node_{2},..., \node_{k} \}$ in coordinate space. With $k$-nearest neighbors, the query point \textbf{p} acquires summarized graph feature $\point^{(i)}_{fp}$ by weighted summation (Eq.~\ref{equation:fp}) of the $k$ node features $\{\node^{(i-1)}_{1}, \node^{(i-1)}_{2},..., \node^{(i-1)}_{k} \}$ $\in$ $\mathbb{R}^{D}$, where the weights are based on the normalized reciprocal distances. Let the distance between the query point and $k$ neighbor nodes be $\{\emph{d}_{1}, \emph{d}_{2},..., \emph{d}_{k} \}$, the propagation can be expressed as:

    \begin{equation}
        \point^{(i)}_{fp} = \frac{\sum_{j=1}^{k} {\node^{(i-1)}_{j}\times\frac{1}{d_{j}}}}{\sum_{l=1}^{k}\frac{1}{d_{l}}}
        \label{equation:fp}
    \end{equation} 

Then, the aggregated feature for point $\point$, $\point^{(i)}_{fp}$ $\in$ $\mathbb{R}^{D}$ is concatenated with the incoming point feature $\point^{(i-1)}$ to create $\point^{(i)}_{concat}$ $\in$ $\mathbb{R}^{2D}$. Finally, an MLP module $\mathcal{F}_{2}:\mathbb{R}^{2D} \rightarrow \mathbb{R}^{D_{next}}$ projects the concatenated point feature to the input dimension of the next layer as $\point^{(i)}$ $\in$ $\mathbb{R}^{D_{next}}$.

\subsection{Implicit Dense Volume Reconstruction}
\label{sec:ipm}

\revised{To acquire dense volume segmentation results, the naive method is to sample all points from the pulmonary tree and group them into many non-overlapping original point clouds for multiple inferences. However, acquiring a dense volume prediction result from repeated point cloud prediction is computationally inefficient~(example in Sec.~\ref{subsec:inference_overhead}) because the graph input remains identical and the point cloud is globally invariant across different inferences.}

\begin{figure}
    \centering
    \includegraphics[width=\linewidth]{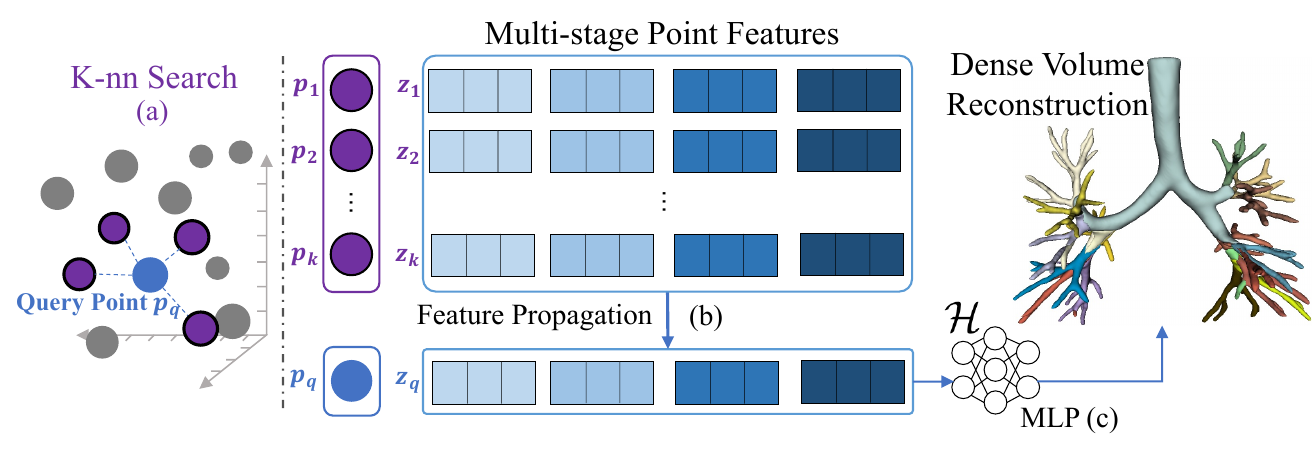}
	\caption{\textbf{Implicit Point Module.} For any query point, the Implicit Point Module consumes multi-stage features from a Point-Graph Network with feature propagation and a neural network to provide a label. }
\label{fig:ipm}
\end{figure}

To improve computation efficiency for dense volume reconstruction, we propose the \emph{Implicit Point Module} in Fig. \ref{fig:ipm} for efficient and arbitrary point inference. \revised{This module effectively interpolates the extracted original point cloud feature from the backbone Point-Graph Network as an feature field to enable fast dense volume reconstruction in a 3-step process. }

First, for arbitrary query point coordinates $\point_{q}=(x_q, y_q, z_q) \in \mathbb{R}^3$ as input, its $k$ nearest-neighbor points $\{\point_{1}, \point_{2}, ..., \point_{k} \}$ in the original point cloud (Fig. \ref{fig:ipm} (a)) are located. Second, for each nearest neighbor point $\point_{k}$, its corresponding features in different stages of the Point-Graph feature fusion are extracted and concatenated to form a multi-stage feature vector $\textbf{z}_{k}=
\{\point^{(0)}_{k} \frown \point^{(1)}_{k} \frown ...  \frown \point^{(i)}_{k}\}$, where $i$ denotes the number of Point-Graph Fusion layers.
A feature propagation (Fig. \ref{fig:ipm} (b)), similar to Eq. \ref{equation:fp}, is performed to aggregate $\{\textbf{z}_{1}, \textbf{z}_{2}, ..., \textbf{z}_{k}\}$ into the feature representation $\textbf{z}_{q}$ for the query point. Finally, an MLP network $\mathcal{H}$ projects the feature $\textbf{z}_{q}$ into a 19-dimensional vector for final classification (Fig. \ref{fig:ipm} (c)). For dense volume segmentation results, we simply sample all foreground points and query through the module for prediction inference.

\revised{As the proposed PGN is regarded as the backbone point-based method, the Implicit Point Module is considered an additional accelerator for dense reconstruction, specializing in generating predictions for arbitrary location based on sparse features from the input point cloud. Therefore, from the perspective of the point-based method, the Implicit Point Module exists is an computation overhead at inference stage that sacrifice an insignificant amount of network~(Fig.~\ref{fig:ipm} $\mathcal{H}$) storage memory, in exchange for largely improved computation efficiency in dense reconstruction.}

\subsection{Model Details}
The IPGN is a customizable pipeline. The initial graph encoder utilizes 11 GAT layers and 128 channels. For ball query in point-to-graph fusion, we set ball radius $r$ = 0.1, the maximum queried point to be 24, and employ a 3-layer GAT network to represent $\mathcal{G}$ in Fig.~\ref{fig:pg_layer}. For feature propagation in both graph-to-point fusion and the Implicit Point Module, we set $k$=3 for the $k$-nearest neighbor search. During the entire Point-Graph Network backbone, we set the intermediate fusion features to be 128-dimensional. 

\subsubsection{Training}
At the point branch, we experimented with two point models, PointNet++~\citep{pointnet++} or Point Transformer~\citep{zhao2021point} as initial feature encoders. For the graph encoder, we apply an 11-layer GAT~\citep{GAT} as the initial feature encoder. Prior to the training of the network, the feature encoders at point and graph branches are independently trained on the corresponding point and graph segmentation tasks and kept frozen. 

The Point-Graph Network and the Implicit Point Module are trained simultaneously. First, we perform a forward pass on the Point-Graph Network, generating multi-stage point and graph features as well as predictions for the $M$ ($6k$) input points and $N$ graph elements. Subsequently, to train the Implicit Point Module, another set of $M'$ ($6k$) foreground points are randomly sampled, and perform forward inference on the Implicit Point Module for $M'$ predictions. After acquiring $M+M'$ point and $N$ graph predictions, we apply the Cross-Entropy loss for training. 

The two point encoder candidates: PointNet++ and Point Transformer, are both trained for 120 epochs with a learning rate of 0.002 while the GAT graph encoder is trained for 240 epochs with a learning rate of 0.02. The IPGN pipeline is trained for 100 epochs with a learning rate of 0.01 and for every 45 epochs, the learning rates are divided by 2. To avoid overfitting, we employ random rotation, shift, and scaling as data augmentation during training. 

\revised{\subsubsection{Inference}
\label{subsec:inference_overhead}

At inference time, the dense volume reconstruction procedure for IPGN involves a single point cloud inference through the backbone Point-Graph Network (PGN) and the inference of the remaining query points using the Implicit Point Module, as detailed in Sec.~\ref{sec:ipm}.

More specifically, to perform dense reconstruction with IPGN, we first execute pre-processing steps including skeletonization (Sec.~\ref{subsec:dataset_preprocess}) and down-sampling to acquire a graph and a point cloud. The graph and the original point cloud are then processed using the Point-Graph Network to obtain multi-stage point-graph features on the point cloud. With these extracted features, we gather all remaining query points belonging to the dense volume foreground and group them into multiple point cloud batches. For example, a dense volume of the airway tree contains approximately $180k$ foreground points, which we group into 4 batches with a batch size of 8 and each point cloud instance containing $6k$ points. Each batch is then iteratively fed into the Implicit Point Module as query points for highly efficient class predictions (Fig.~\ref{fig:ipm}).

Comparatively, for an arbitrary point-based backbone, such as the proposed Point-Graph Network or Point Transformer~\citep{zhao2021point}, dense volume acquisition involves multiple full-model inferences on many point clouds that collectively assemble the dense volume. However, this method is computationally inefficient, as shown in Sec.~\ref{subsec:Imp_mod} and Table~\ref{tab:time-time}.}
\section{Experiments}
\label{sec:experiments}

\begin{table*}[h!]
\centering
\resizebox{\textwidth}{!}{%
\begin{tabular}{l|ccc|ccc|ccc|ccc}
\hline
\multicolumn{1}{c|}{\multirow{3}{*}{Methods}} & \multicolumn{3}{c|}{Representation} & \multicolumn{3}{c|}{Airway} & \multicolumn{3}{c|}{Artery} & \multicolumn{3}{c}{Vein} \\
\multicolumn{1}{c|}{} & \multirow{2}{*}{Voxel} & \multirow{2}{*}{Point} & \multirow{2}{*}{Graph} & \multicolumn{1}{c|}{\multirow{2}{*}{Voxel (\%)}} & \multicolumn{2}{c|}{Graph / KP$^{*}$ (\%)} & \multicolumn{1}{c|}{\multirow{2}{*}{Voxel (\%)}} & \multicolumn{2}{c|}{Graph / KP$^{*}$ (\%)} & \multicolumn{1}{c|}{\multirow{2}{*}{Voxel (\%)}} & \multicolumn{2}{c}{Graph / KP$^{*}$ (\%)} \\
\multicolumn{1}{c|}{} &  &  &  & \multicolumn{1}{c|}{} & \multicolumn{1}{c|}{Node} & Edge & \multicolumn{1}{c|}{} & \multicolumn{1}{c|}{Node} & Edge & \multicolumn{1}{c|}{} & \multicolumn{1}{c|}{Node} & Edge \\ \hline
 {\color[HTML]{34696D} \textit{Voxel / Point}} &  &  &  &  &  &  &  &  &  &  &  & \\
3D-Unet (down-sampled)~\citep{unet3d} & \checkmark &  &  & 58.5 &  61.3 & 62.3 & 61.4 & 62.0 & 63.1 & 54.0 & 55.2 & 55.0 \\
3D-Unet (down-sampled) + KP~\citep{tsinghua} & \checkmark &  &  & 56.9 & 62.2 & 61.9 & 61.3 & 62.1 & 63.1 & 55.0 & 55.4 & 54.8 \\
3D-Unet (sliding-window)~\citep{unet3d} & \checkmark &  &  & 39.8 & 40.3 & 41.2 & 22.5 & 23.5 & 24.1 & 28.7 & 29.3 & 28.7 \\
PointNet~\citep{Qi2016PointNetDL} &  & \checkmark &  & 79.1 & 80.5& 79.6 & 80.0 & 80.5 & 80.4 & 70.8 & 70.8 & 70.1 \\
PointNet++~\citep{pointnet++} &  & \checkmark &  & 82.9 & 84.0 & 83.5 & 82.5 & 82.8 & 83.5 & 75.7 & 75.7 & 75.6 \\ 
Point Transformer~\citep{zhao2021point} &  & \checkmark &  & 87.8 & 87.2 & 87.4 & 86.5 & 87.2 & 87.4 & 77.7 & 77.2 & 78.0 \\\hline
{\color[HTML]{34696D} \textit{Graph}} &  &  &  &  &  &  &  &  &  &  &  & \\
GCN~\citep{GCN} &  &  & \checkmark & 82.8 & 85.5 & 83.0 & 79.6 & 82.1 & 80.4 & 73.7 & 73.5 & 70.4 \\
GIN~\citep{GIN} &  &  & \checkmark & 84.6 & 88.8 & 85.6 & 81.4 & 84.9 &  82.8 & 74.0 & 74.3 & 70.9 \\
GraphSage~\citep{GraphSage} &  &  & \checkmark & 86.0 & 91.7 & 88.0 & 82.4 & 86.7 & 84.5 & 75.9 & 79.1 & 75.0 \\
HyperGraph~\citep{TNN} &  &  & \checkmark & 86.0 & 93.0 & 88.8 & 82.2 &  89.0 & 85.7 & 75.7 & 79.4 & 75.8 \\
HyperGraph w/ Handcrafted Feat~\citep{TNN} &  &  & \checkmark & 85.8 & 92.8 & 88.7 & 81.8 & 88.6 & 85.2 & 75.7& 79.5 & 75.4 \\
GAT~\citep{GAT} &  &  & \checkmark & 86.3 & 92.7 & 89.1 & 83.4 & 88.7 & 86.3 & 76.0 & 79.0 & 75.3 \\
GAT w/ Handcrafted Feat~\citep{TNN} &  &  & \checkmark & 86.3 & 92.5 & 88.8 & 83.7 & 88.4 & 86.2 & 76.1 & 78.1 & 74.1 \\ \hline
{\color[HTML]{34696D} \textit{Point + Graph}} &  &  &  &  &  &  &  &  &  &  &  & \\
GAT w/ PointNet++ Feat~\citep{pointnet++} &  & \checkmark & \checkmark & 86.7 & 94.6 & 91.8 & 83.8 & 90.5 & 88.6 & 76.3 & 80.4 & 77.0 \\
PGN (PointNet++~\citep{pointnet++}) &  & \checkmark & \checkmark & 87.7 & \textbf{95.1} & 92.3 & 86.8 & \textbf{92.9} & \textbf{89.9} & 77.4 & \textbf{82.8} & \textbf{79.4} \\
IPGN (PointNet++~\citep{pointnet++})&  & \checkmark & \checkmark & 87.5 & \textbf{95.1} & 92.3 & 86.6 & \textbf{92.9} & \textbf{89.9} & 77.5 & \textbf{82.8} & \textbf{79.4} \\ PGN (Point Transformer~\citep{zhao2021point}) &  & \checkmark & \checkmark & 88.4 & 95.0 & \textbf{92.7} & \textbf{87.2} & 92.6 & \textbf{89.9} & \textbf{78.6} & 82.3 & 78.8 \\
IPGN (Point Transformer~\citep{zhao2021point}) &  & \checkmark & \checkmark & \textbf{88.5} & 95.0 & \textbf{92.7} & \textbf{87.2} & 92.6 & \textbf{89.9} & 78.3 & 82.3 & 78.8 \\ \hline
\end{tabular}%
}
\caption{\textbf{Model Performance in Dice Score at Voxel and Graph Level on the PTL dataset.} Baseline methods using different feature representations are presented to compare against the proposed methods on the Pulmonary Tree Labeling (PTL) dataset. PGN=Point-Graph Network without implicit fields; IPGN=Implicit Point Graph Network; KP=key point. *:~Graph-level predictions from Voxel/Point Networks are acquired by inferencing on the node and edge locations. } 
\label{tab:main_table}
\end{table*}

\subsection{Experiment Setting}
\label{subsec:exp_setting}
By default, all experiments use dense volume as initial input. While CNN is natural for manipulating dense volume, we pre-process the dense volume to point clouds and skeleton graphs for point and graph experiments. We present the experiment metrics at the voxel-level, representing overall performance, and graph-level, representing performance at key locations in the structure. Although voxel and point-based methods do not inherently provide evaluations at the graph-level, we acquire the predictions at these key coordinates to provide the evaluation at key points for voxel/point methods. Moreover, voxel-level dense volume performance evaluation can be achieved using CNN methods, point-based methods with repeated inferences and graph-based models after post-processing (Sec.~\ref{subsec:graph_setting}). Therefore, the evaluations in voxel-level, and graph-level across CNN, point, and graph methods are consistent and fair. 

\subsubsection{CNNs}
\label{subsec:cnn_setting}
In CNN experiments, given a 3D input, the task involves providing semantic segmentation prediction for the pulmonary tree in the image and all CNN methods apply a combination of dice loss and the cross-entropy loss for training. During testing, the image background is excluded from the metric computation.

Among the CNN experiments, we employ 3D-Unet~\citep{unet3d} as the basic setup. Based on the 3D-Unet, we also implement a multi-task key-point regression method~\citep{tsinghua}, abbreviated as "3D-Unet + KP". More specifically, an additional regression prediction head predicts a heatmap representing how likely is the location of a graph node as a key point. In these two CNN-based experiments, data are down-sampled to $96\times96\times96$ due to limited GPU memory during training and validation. During testing, the input is down-sampled for inference and re-scaled to the original dimension $N\times512\times512$ for evaluation.

To address the issue of high memory usage and information loss by compromised resolution, we apply a sliding window approach in another CNN experiment, in which local 3D patches with dimension $96\times96\times96$ from the original image are used for training and validation purposes. During testing, the predictions obtained from the sliding-window technique were assembled back onto the original image for evaluation.

\subsubsection{Point Clouds}
\label{subsec:point_setting}
Point-based experiments involve treating a set of tubular voxels of a pulmonary tree as a point cloud for sparse representation for modeling. At the output, point-based model provides per-point classification prediction.

During training and validation of the experiments, we randomly sampled $6k$ foreground elements as point cloud input for this dataset, as sampling more than $6k$ points consumes more resources but does not increase performance. During testing, all foreground points are sampled, randomly permuted, and grouped into multiple point clouds. Then each point cloud containing $6k$ points is iteratively fed into the model for evaluation. Consequently, the inference processes provide a prediction for all foreground elements as dense volume results. PointNet~\citep{Qi2016PointNetDL}, PointNet++~\citep{pointnet++} and Point Transformer~\citep{zhao2021point} are tested as baselines.

\subsubsection{Graphs}
\label{subsec:graph_setting}

Graph experiments utilize the skeleton-graph generated from dense volume~\citep{vesselvio, LEE1994} as the graph structure, evaluating networks' ability to recognize key points and structural connections within the skeleton tree. These abilities are respectively represented by dice scores on graph nodes and edges.

In these experiments, we leverage multiple graph neural networks (GNNs) such as GAT~\citep{GAT}, GraphSage~\citep{GraphSage}, and a graph network with pre-trained point-based features as input. \revised{To identify an optimal configuration for graph-based networks, we conducted experiments using different GAT layers and feature embedding sizes (see Table~\ref{table:parameter}) on the PTL dataset. We concluded that a configuration of 14 layers with 512-dimensional features is optimal, as it prevents under-fitting while maintaining high efficiency. Deeper GNNs fail to produce higher performance due to the known oversmoothing effect~\citep{Cai2020ANO_oversmoothing} in deep GNNs, and larger feature embeddings also did not yield better results. Consequently, we applied a 14-layer network for each graph baseline.} After acquiring node features, features from the source node and destination node are interpolated by averaging for edge features before an MLP network projects all features to 19 dimensions for final classification.

Unlike voxel/point methods, Graph Neural Networks that perform learning on the skeleton graph provide evaluation at the graph level but not at the voxel level. To address this, we implemented a post-processing technique to dilate graph-based predictions to dense volume predictions. Specifically, for any given voxel, the algorithm assigns the label of the nearest graph element as its own label. Therefore, all graph baselines also provide voxel-level prediction metrics.

\subsubsection{Ours}
\label{subsec:proposed_setting}
As discussed in depth in Sec.~\ref{sec:method}, we perform experiments combining point learning with graph learning. For the proposed Point-Graph network, the input and output setups for the point branch and graph branch are identical to those of point experiments and graph experiments. To speed up dense reconstruction, we incorporate the Implicit Point Module for point-based prediction, evaluated at voxel-level.

\subsection{Model Performance Comparison}
In this section, we perform a comparative analysis based on Table \ref{tab:main_table}. The statistics presented in this section are the average performance over the 3 pulmonary structures. As previously discussed, we will treat graph elements as key locations in the tree structures. Therefore, ``graph-level" is synonymous to ``key point" in evaluation.

For voxel-based methods, the classic 3D-Unet~\citep{unet3d} is used to implement multiple baselines. Two 3D-Unet~\citep{unet3d, tsinghua} methods based on down-sampled input yield unsatisfying dice metrics at voxel-level, indicating that training CNN methods on reduced resolution leads to inferior modeling. As an attempt to train on the original resolution without memory restriction, the 3D-Unet~\citep{unet3d} experiment with sliding-window strategy reports the poorest performances. Among point-based methods, the Point Transformer~\citep{zhao2021point} achieves the overall best voxel-level dice performances over PointNet++~\citep{pointnet++} and PointNet~\citep{Qi2016PointNetDL}, reaching an average of 84\% in dice, which largely exceeds performances by any CNN methods. As for the graph-level results with voxel/point-based methods, namely the voxels/points performance at graph elements' locations, they are closely aligned with the respective voxel-level results, as there is no particular focus on key locations in these methods.

Among GNN methods with graph-only context, GAT~\citep{GAT} model, with minimal tuning, outperforms all previous baselines with a considerable margin, reaching 86.8\% in dice score for node, displaying the advantage of tailored graph learning on the skeleton, focusing on key locations. In terms of graph-based voxel-level results derived from post-processing(Sec.~\ref{subsec:dataset_preprocess}), the GAT~\citep{GAT} model also achieves the best outcome, reaching 81.9\% on average. Given that GNN baselines are highly light-weight and lack shape context, the reconstruction results are auspicious, indicating that the skeleton-graph already provides salient information.

In addition to graph modeling with 3D coordinate features, the performance of the GAT model, and a hypergraph~\citep{TNN} model with handcrafted features~\citep{TNN} are presented. \revised{The handcrafted feature for the graph nodes are carefully curated~(Details in Sec.~\ref{sec:ablation_feature_input}) to best resemble the original description~\citep{TNN}, including structural, positional and morphological features.} Compared to applying coordinate features as input, GAT with handcrafted features suffers from a slight drop in overall performance. Similarly, applying handcrafted features to the hypergraph~\citep{feng2019hypergraph} model 
also translates to a performance drop. These results suggest that in graph settings, tailor-designed features do not provide more valuable information than 3D coordinates. More insights with experiments are provided in Sec.~\ref{sec:ablation_feature_input}.

For settings with both graph and point context, GAT with pre-trained PointNet++ feature has the lower performance compared to the proposed methods. Nevertheless, it still outperforms any baseline that does not involve point-graph fusion-feature by a considerable margin. Such a gap in performance indicates that the integration of point-context into graph learning is beneficial. For the proposed Point-Graph Network and IPGN, their performances beat all baselines in metrics at both levels, displaying superiority in point-graph fusion learning.
 
\subsection{Dense Volume Reconstruction}
\label{subsec:Imp_mod}

\begin{table*}[tb]
\centering
\resizebox{\textwidth}{!}{%
\begin{tabular}{cc|cccc|cccc|cccc}
\hline
\multirow{2}{*}{Methods} & \multirow{2}{*}{Model} & \multicolumn{4}{c|}{Airway} & \multicolumn{4}{c|}{Artery} & \multicolumn{4}{c}{Vein} \\
 &  & Time~(s) & Voxel~(\%) & Node~(\%) & Edge~(\%) & Time~(s) & Voxel~(\%) & Node~(\%) & Edge~(\%) & Time~(s) & Voxel~(\%) & Node~(\%) & Edge~(\%)\\ \hline
{\color[HTML]{34696D} \textit{Voxel}} & 3D-Unet~\citep{unet3d} (downsampled) & 8.16 & 58.5 & 61.3 & 62.3 & 8.02 & 61.4 & 62.0 & 63.1 &  7.46 & 54.0 & 55.2 & 55.0 \\
{\color[HTML]{34696D} \textit{Voxel}} & 3D-Unet~\citep{unet3d} (sliding-window) & 9.43 & 39.8 & 40.3 & 41.2 & 11.88 & 22.5 & 23.5 & 24.1 & 12.44 & 28.7 & 29.3 & 28.7\\
{\color[HTML]{34696D} \textit{Point}} & PointNet++~\citep{pointnet++} & 4.51 & 82.9 & 84.0 & 83.5 & 9.17s & 82.5 & 82.8 & 83.5 & 9.65 & 75.7 & 75.7 & 75.6\\ 
{\color[HTML]{34696D} \textit{Point}} & Point Transformer~\citep{zhao2021point} &10.97 & 87.8 & 87.2 & 87.4 & 21.94 & 86.5 & 87.2 & 87.4 & 23.25 & 77.7 & 77.2 & 78.0\\
{\color[HTML]{34696D} \textit{Graph}} & GAT~\citep{GAT} & \textbf{1.15} & 86.3 & 92.7 & 89.1 & 2.05 & 83.4 & 88.7 & 86.3 &2.24 & 76.0 & 79.0 & 75.3 \\
\hline
{\color[HTML]{34696D} \textit{Point + Graph}} & PGN (PointNet++~\citep{pointnet++}) & 5.65 & 87.7 & 95.1 & 92.3 & 12.40 & 86.8 & \textbf{92.9} & \textbf{89.9} & 13.16 & 77.4 & \textbf{82.8} & \textbf{79.4}\\
{\color[HTML]{34696D} \textit{Point + Graph}} & IPGN (PointNet++~\citep{pointnet++}) & 1.30 & 87.5 & 95.1 & 92.3 & \textbf{1.49} & 86.6 & \textbf{92.9} & \textbf{89.9} & \textbf{1.57} & 77.5 & \textbf{82.8}& \textbf{79.4}\\
{\color[HTML]{34696D} \textit{Point + Graph}} & PGN (Point Transformer~\citep{zhao2021point}) & 12.10 & 88.4 & \textbf{95.0} & \textbf{92.7} & 24.28 & \textbf{87.2} & 92.6 & \textbf{89.9} & 25.81 & \textbf{78.6} & 82.3 & 78.8\\
{\color[HTML]{34696D} \textit{Point + Graph}} & IPGN (Point Transformer~\citep{zhao2021point}) & 2.32 & \textbf{88.5} & \textbf{95.0} & \textbf{92.7} & 2.29 & \textbf{87.2} & 92.6 & \textbf{89.9} & 2.39 & 78.3 & 82.3 & 78.8\\\hline
\end{tabular}%
}
\caption{\textbf{Inference Speed and Segmentation Metrics \revised{for Dense Volume Reconstruction}.} This table compares the dense volume segmentation test time and dice score across voxel, point, graph-based, and point-graph fusion methods. The test times are measured in seconds while dice score presents segmentation quality.}
\label{tab:time-time}
\end{table*}

In this subsection, we focus on the Implicit Point Module (Fig. \ref{fig:ipm}), which aims to enhance the efficiency of labeled dense volume reconstruction (Fig. \ref{fig:architecture} (f)). As discussed previously in Sec. \ref{sec:introduction}, the integration of the implicit module is necessary because high-performing point models like Point Transformer~\citep{zhao2021point} could still be computationally expensive at test time: with repeated inferences for dense prediction, its computation cost grows in cubicle, similar to CNN. \revised{Experiments are completed to compare the efficiency using the proposed IPGN against only using the backbone Point-Graph Network as well as various baseline methods.} The reconstruction efficiency is represented as the average run-time in seconds, for dense volume prediction across the PTL test set.

Regarding volume reconstruction efficiency, the test time of convolutional methods, graph-based method, point-based method, and our methods are ducomented. For all setups\footnote{To make it easily reproducible, all the speed measurements were conducted on a free Google Colab T4 instance---CPU: Intel(R) Xeon(R) CPU @ 2.00GHz, GPU: NVIDIA Tesla T4 16Gb, memory: 12Gb.}, we present the inference time and quality for dense volume segmentation in Table~\ref{tab:time-time}. Apart from model inference costs, test time is composed of various operations in different setups. For CNN with down-sampled input, test time includes down-sampling and up-sampling operations. For graph baselines, time measurement takes post-processing~(Sec.~\ref{subsec:graph_setting}) into account. \revised{For point-based networks, dense reconstruction time consists of multiple batched point cloud inferences.} Finally, test time for IPGN includes a forward pass on the Point-Graph Network and the subsequent Implicit Point Module inference on all remaining points. 

The results demonstrated the superiority of the IPGN in dense volume reconstruction. \revised{Qualitatively, the IPGN, with predictions from the Implicit Point Module rather than the backbone PGN, achieve nearly identical performance as the Point-Graph Network.} In terms of efficiency, IPGN only requires $1.5$ seconds on average with PointNet++~\citep{pointnet++} and $2.3$ seconds with Point Transformer~\citep{zhao2021point} as encoder. \revised{This reconstruction efficiency of IPGN is largely attributed to the Implicit Point Module. Let us explain the computational analysis of PGN and IPGN using dense reconstruction of the airway as an example. 
As shown in Table~\ref{tab:time-time}, the average inference time for PGN is 5.65 seconds, which can be broken down into an average of 3.95 batches per airway, with each batch containing 8 point clouds of $6k$ points, taking 1.43 seconds. In contrast, IPGN only requires running a single batch of $6k$ points through the PGN, taking 1.03 seconds, followed by dense volume reconstruction of all query points in 0.27 seconds, totaling just 1.30 seconds. Thus, by introducing the Implicit Point Module, we achieve a reduction of over 4 seconds with insignificant computational overhead. For more complex cases, such as inputs with more points (arteries and veins) and computationally intensive backbones (Point Transformer), the time savings with IPGN are even more significant.}

Although the graph-based method records a similar time cost, it fails to generate quality results. Further, all other methods are relatively inefficient. Therefore, the proposed IPGN pipeline is the all-around optimal solution for dense volume segmentation for pulmonary structures.

These findings highlight the practical utility of the proposed module, as it allows for fast and efficient point inference without compromising quality. The Implicit Point Module serves as a valuable contribution in overcoming the computational challenges associated with analyzing the pulmonary tree, enabling rapid decision-making in clinical settings.

\begin{figure*}[!htb]
    \centering
    \includegraphics[width=\linewidth]{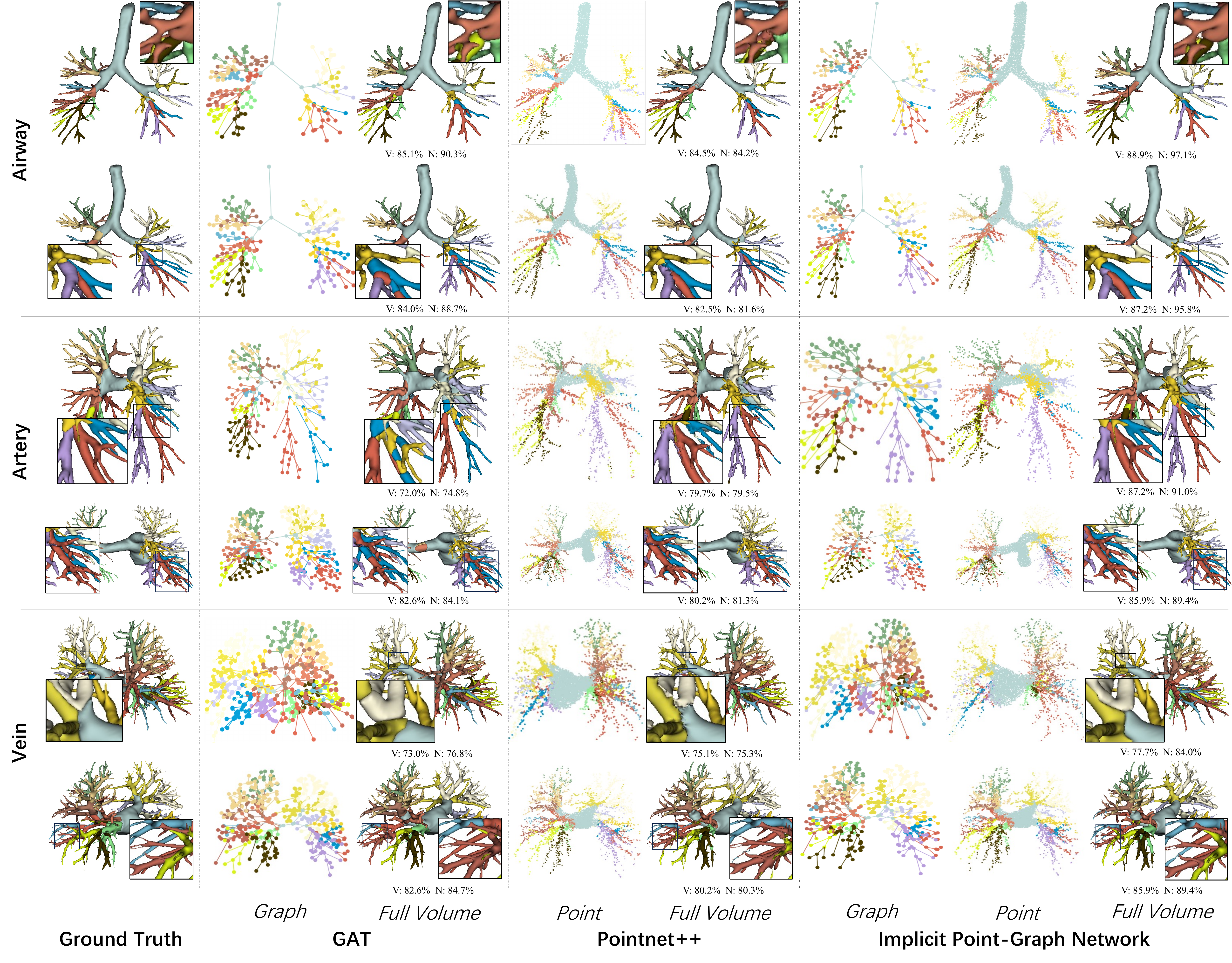}
	\caption{\textbf{Visualization of Segmentation Results.} This figure displays segmentation results using GAT~\citep{GAT} w/ post-processing~(Sec.~\ref{subsec:graph_setting}), PointNet++~\citep{pointnet++} and IPGN (with GAT and PointNet++ backbones) methods along with the ground truth. The initial-form predictions (graph or point cloud) before dense volume prediction are also presented. Below each dense volume reconstruction, we show voxel~(V) dice and node~(N) dice results.} 
    \label{fig:qualitative}
\end{figure*}

\subsection{Qualitative Analysis with Visualization}
\label{subsec:qualitative}

In the qualitative analysis section, we employ 2 concrete cases each from the airway, artery, and vein tree to demonstrate the efficacy of our method. Fig. \ref{fig:qualitative} (a-f) showcases the results of full pulmonary tree segmentation using graph-only method, point-only method, and the Implicit Graph-Point Network, against the ground truth image. 

The graph-based predictions from Fig. \ref{fig:qualitative} (b-d) reveal instances of incorrect graph predictions leading to anomalous outcomes, where multiple class labels appear in the same branch, thereby inducing inconsistent branch predictions. For point-based method, due to its sparse representation in individual forward passes, point predictions often lose details and disrupt the borders between branches (Fig. \ref{fig:qualitative} (a,c,e)), resulting in volumes that lack smoothness and quality. 

Conversely, the proposed IPGN offers various advantages. Firstly, our method accurately classifies distal branches, illustrated by dense volume prediction in Fig. \ref{fig:qualitative} (d,f). Additionally, our method effectively defines clear and uninterrupted borders between the child branches at bifurcation points. This is especially valuable when a parent node branches into multiple child branches belonging to different classes, as demonstrated in Fig. \ref{fig:qualitative} (b,e). By integrating graph and point context into the modeling process, our method enhances the segmentation capabilities at bifurcation points and distal branches, ultimately producing smoother and more uniform branch predictions.

\subsection{Ablation Experiments}
\label{subsec:exp_Abl}

\subsubsection{Feature for Graph Modeling}
\label{sec:ablation_feature_input}

In this section, we examine the effect of different input features for the skeleton graph node. The candidates are 3D coordinates feature and handcrafted feature. As coordinate feature is simply the $\{x,y,z\}$ coordinates, handcrafted feature follows a similar feature design from \cite{TNN}. 

\revised{In the experiment setup of~\cite{TNN}, a graph node represents an airway tree branch in volumetric space, with features including structural, positional, and morphological attributes derived from the tree branch. In our proposed pipeline, the volume is abstracted as a skeleton graph, with graph nodes representing bifurcation points or points along the centerline of branches. We slightly modified the original feature design to fit our skeleton graph setup, described as follows. Our version of the feature design can be adapted to airway, artery, and vein trees, instead of being limited to airway trees.

For structural features such as generation, number of children and siblings, these can be derived from the skeleton graph by treating the extra-pulmonary structure (class=19) as a starting point (first generation) and iteratively calculating the structural information for each branching point until reaching leaf nodes.

For positional features, the feature design indicates that the spatial locations of branches can be used in various ways, such as using an anchor location to establish coordinate systems. All positional features can be acquired for the graph because each node is represented by its 3D coordinates. For the airway tree, we follow the original design by treating the end of the trachea as the anchor. For the artery and vein trees, we use the average coordinates of all nodes representing extra-pulmonary structures as anchors. Positional features also include creating a bounding box for tree structures in the left and right lung to establish another coordinate system. Additionally, branches are sorted, and their sorted orders and angles are also used as positional features. In our implementation, rather than each branch, we rank each graph node in the three spatial dimensions and calculate the angles of edges that connect each node.

For morphological features, we use the length of branches and the projected lengths of branches in three axes as features. This can be implemented at the node level by calculating the corresponding total lengths of the edges that connect to a node.}

In our experiments, we use a GAT~\citep{GAT} and a 5-layer MLP network to evaluate the impact of coordinate features and handcrafted features, reported in Table \ref{tab:ablation_feature_mlp}. Based on the MLP experiment, handcrafted feature achieves 78.4\% dice results on average while raw coordinate only reports 71.5\%. Conversely, in the experiment with GAT~\citep{GAT}, applying the coordinate feature achieves overall best performances, reaching 86.8\% while the handcrafted feature trails behind.

The results indicate that the handcrafted feature provides more information and produces better performance in an MLP setting, where the neural network simply learns a non-linear projection without any global or tree topology context. Once the connections between graph nodes are established through edge in graph modeling, the handcrafted feature completely loses its advantages over the raw coordinate feature, which implies that the handcrafted feature could be learned during graph learning and the learned graph-based feature is more beneficial to graph segmentation quality.

\subsubsection{Input Selection For Implicit Point Module}

\begin{table}
\centering
\resizebox{\linewidth}{!}{%
\begin{tabular}{c|cc|cc|cc|cc}
\hline
\multirow{2}{*}{Method} & \multicolumn{2}{c|}{Features} & \multicolumn{2}{c|}{Airway} & \multicolumn{2}{c|}{Artery} & \multicolumn{2}{c}{Vein} \\
 & Handcrafted & Raw Coord & Node & \multicolumn{1}{c|}{Edge} & Node  & \multicolumn{1}{c|}{Edge } & Node  & Edge  \\ \hline
\multirow{3}{*}{MLP} & \checkmark &  & \textbf{80.8} & \textbf{81.2} & 78.1 & \textbf{77.6} & 74.2 & 74.3 \\
 &  & \checkmark & 69.2 & 69.7 & 72.4 & 72.2 & 71.9 & 72.0 \\
 & \checkmark & \checkmark & 80.7 & 80.9 & \textbf{78.2} & \textbf{77.6} & \textbf{75.5} & \textbf{75.3} \\ \hline
\multirow{3}{*}{GAT~\citep{GAT}} & \checkmark &  & 92.0 & 89.5 & 88.0 & 85.6 & 77.8 & 73.4 \\
 &  & \checkmark & \textbf{92.7} & 89.1 & \textbf{88.7} & \textbf{86.3} & \textbf{79.0} & \textbf{75.3} \\
 & \checkmark & \checkmark & 92.2 & \textbf{89.2} & 88.2 & 86.0 & 78.5 & 75.1 \\ \hline
\end{tabular}%
}
\caption{\textbf{Impact of Input for Graph Modeling.} We show the results using MLP and GAT~\citep{GAT} backbones on different feature inputs.}
\label{tab:ablation_feature_mlp}
\end{table}

\begin{table}
\centering
\resizebox{\linewidth}{!}{%
\begin{tabular}{l|c|c|c|c}
\hline
 & $\textbf{P}^{(3)}$ & $\textbf{P}^{(2)}\frown \textbf{P}^{(3)}$ & $\textbf{P}^{(1)}\frown \textbf{P}^{(2)}\frown \textbf{P}^{(3)}$ & $\textbf{P}^{(0)}\frown \textbf{P}^{(1)}\frown \textbf{P}^{(2)}\frown \textbf{P}^{(3)}$ \\ 
 \hline
Airway & 88.1 & 88.2 & 88.4 & \textbf{88.5} \\
{Artery} & 86.7 & 87.0 & \textbf{87.0} & \textbf{87.2} \\
{Vein} & 77.9 & 77.9 & \textbf{78.1} & \textbf{78.3} \\ \hline
\end{tabular}%
}
\caption{\textbf{Multi-stage Point Feature Input for Implicit Point Module.} This table presents the dice results using different combinations of concatenated multi-stage features from the backbone network as input to the Point Implicit Module. (Point Encoder: Point Transformer~\citep{zhao2021point}; Graph Encoder: GAT~\citep{GAT})}
\label{table:imp_feature_select}
\end{table}

In this section, we present the results of an ablation study where we explore the impact of multi-stage features using different layers of the intermediate feature vector as input to the implicit module.

In the experiment, we initially utilized the feature output of the final Point-Graph Fusion layer as the sole input to the implicit module. Motivated by the design of DenseNet~\citep{Huang2016DenselyCC}, we experimented with adding feature outputs from shallower intermediate blocks, forming multi-stage features, and report the performance with each feature addition until the initial point feature.
Table \ref{table:imp_feature_select} demonstrates that combining multi-stage features, along with the initial point (PointNet++~\citep{pointnet++}) feature, enhances the predictive capabilities of our model, contributing to better performance. Finally, the best-performing configuration we observed involved using all available features, yielding results on par with the full modeling approach.

\revised{\subsection{Robustness Test on Corrupted Data}

As the proposed pipeline is intended to be deployed in diverse environments, high-quality pre-extractions (binary segmentation) of the 3D tree structures from CT scans could be challenging due to the complex tubular shapes, especially at distal endpoints, where thin tubular branches in the resulting volume could be disconnected. However, we believe pulmonary tree segmentation and pulmonary tree labeling are distinct yet important issues, as standard segmentation methods might struggle with pulmonary tree labeling, as shown in Table~\ref{tab:main_table} and Table~\ref{tab:time-time} with the 3D Unet results.

To test the robustness of our model under poorly constructed trees, we performed additional experiments on a corrupted version of the PTL dataset, which contains pulmonary trees with disconnections~\citep{weng2023topology}. The results, presented in Table 6 of the manuscript, indicate that while both Point Transformer and IPGN are robust to volume-level corruption, the proposed method is relatively more robust compared to point-based methods. Overall, the corruption in branches does not pose a major threat to the overall segmentation of the target structures. }

\begin{table}
\centering
\resizebox{\linewidth}{!}{%
\begin{tabular}{l|lllc}
\hline
\multicolumn{1}{c|}{Methods} & \multicolumn{1}{c}{Airway} & \multicolumn{1}{c}{Artery} & \multicolumn{1}{c}{Vein} & \multicolumn{1}{l}{Average $\Delta$} \\ \hline
Point Transformer~\citep{zhao2021point} & 87.8/ \red{85.4} & 86.5/\red{85.0} & 77.7/ \red{74.1} & \red{2.5}$\downarrow$ \\
IPGN (Point Transformer~\citep{zhao2021point}) & 88.5/\red{88.0} & 87.2/\red{86.1} & 78.3/\red{76.2} & \red{1.2}$\downarrow$ \\ \hline
\end{tabular}%
}
\caption{\revised{\textbf{Model Performance on the Original / Corrupted Data.} We present experiment results on the original and corrupted version~\citep{weng2023topology} of the PTL dataset for Point Transformer~\citep{zhao2021point} and the proposed IPGN with voxel-level dice score~(\%) as metric.}}
\label{table:robustness}
\end{table}

\revised{\subsection{Model Complexity}

\begin{table}
\centering
\resizebox{\linewidth}{!}{%
\begin{tabular}{lcccccc}
\hline
\multicolumn{1}{c}{} & \multicolumn{2}{c}{Configuration} &  &  &  &  \\
\multicolumn{1}{c}{\multirow{-2}{*}{Methods}} & Feature & Layer & \multirow{-2}{*}{Airway} & \multirow{-2}{*}{Artery} & \multirow{-2}{*}{Vein} & \multirow{-2}{*}{\begin{tabular}[c]{@{}c@{}}Parameters
\end{tabular}} \\
\hline

 & 512 & 4 & 82.9 & 82.5 & 75.7 & 1.0M \\
\multirow{-2}{*}{PointNet++~\citep{pointnet++}} & 1024 & 6 & 83.4 & 82.8 & 76.8 & 3.9M \\ \cline{2-7} 
 & 512 & 4 & 87.8 & 86.5 & 77.7 & 4.5M \\
\multirow{-2}{*}{Point Transformer~\citep{zhao2021point}} & 1024 & 6 & 88.1 & 86.7 & 78.0 & 12.5M \\ \hline

 & 128 & 11 & 86.0 & 81.4 & 74.0 & 1.3M \\ \cline{2-7} 
 &  & 11 & 86.2 & 81.8 & 74.4 & 1.8M \\
 & \multirow{-2}{*}{256} & 14 & 86.2 & 82.9 & 75.4 & 2.2M \\ \cline{2-7} 
 &  & 11 & 86.0 & 82.5 & 74.8 & 3.9M \\
 &  & 14 & 86.3 & 83.4 & 76.0 & 5.5M \\
 &  & 20 & 87.0 & 84.2 & 76.1 & 6.1M \\
 & \multirow{-4}{*}{512} & 25 & 86.6 & 83.2 & 75.7 & 8.6M \\ \cline{2-7} 
 &  & 14 & 87.1 & 82.9 & 76.3 & 12.3M \\
 &  & 20 & 86.9 & 82.0 & 76.0 & 13.7M \\
\multirow{-10}{*}{GAT~\citep{GAT}} & \multirow{-3}{*}{1024} & 25 & 85.5 & 83.0 & 75.3 & 15.5M \\ \hline

PGN (PointNet++~\citep{pointnet++}) & \multicolumn{2}{c}{} & 87.7 & 86.8 & 77.4 & 3.6M \\
IPGN (PointNet++~\citep{pointnet++}) & \multicolumn{2}{c}{} & 87.5 & 86.6 & 77.5 & 3.7M \\
PGN (Point Transformer~\citep{zhao2021point}) & \multicolumn{2}{c}{} & 88.4 & \bf 87.2 & \bf 78.6 & 7.2M \\
IPGN (Point Transformer~\citep{zhao2021point}) & \multicolumn{2}{c}{\multirow{-4}{*}{\textit{N/A}}} & \bf 88.5 & \bf 87.2 & 78.3 & 7.3M \\ \hline
\end{tabular}%
}
\caption{\revised{\textbf{Model Complexity.} This table highlights the parameter counts for key competitive methods. The performance is represented as voxel-level dice (\%).}}
\label{table:parameter}
\end{table}

To demonstrate that the performance improvement of PGN/IPGN is not merely due to an increase in parameter count, we scaled up the size of both graph-based and point-based models to incorporate model complexity into the comparison. We recorded the model parameter count for the graph-based methods, point-based methods, and the proposed methods in Table~\ref{table:parameter}.

For the graph-based models, we evaluated the highest-performing baseline, GAT~\citep{GAT}, under different configurations. As the GAT layer grows from 11 to 25 layers and the maximum feature embedding grows from 256 to 1024, the parameter count increases considerably from $1.8M$ to $15.5M$, largely exceeding our methods. However, increasing the parameter count of the GAT model does not contribute to significant performance gains, which aligns with the known over-smoothing effect in deep graph modeling~\citep{Cai2020ANO_oversmoothing}.Similarly for point-based methodsy, while increasing the model size may improve performance, it is far less effective than the enhancements provided by Point-Graph Fusion.

Our PGN/IPGN combines the advantages of point representation and graph representation. For example, among the $3.6M$ parameters of the PGN (PointNet++), $1.0M$ parameters are from PointNet++, $1.3M$ parameters are from GAT networks, and 1.3M parameters are from Point-Graph feature fusion layers (Sec.~\ref{sec:pgf}). As demonstrated, the performance improvement from this combination is highly rewarding. 

Finally, while IPGN adds a small number of parameters ($0.1M$) compared to PGN, it results in a significant efficiency improvement (Table~\ref{tab:time-time}).}

\section{Extended Application: Reconstruction of Pulmonary Segments without Retraining}
\label{sec:extended-app}

\begin{figure}
    \centering
    \includegraphics[width=\linewidth]{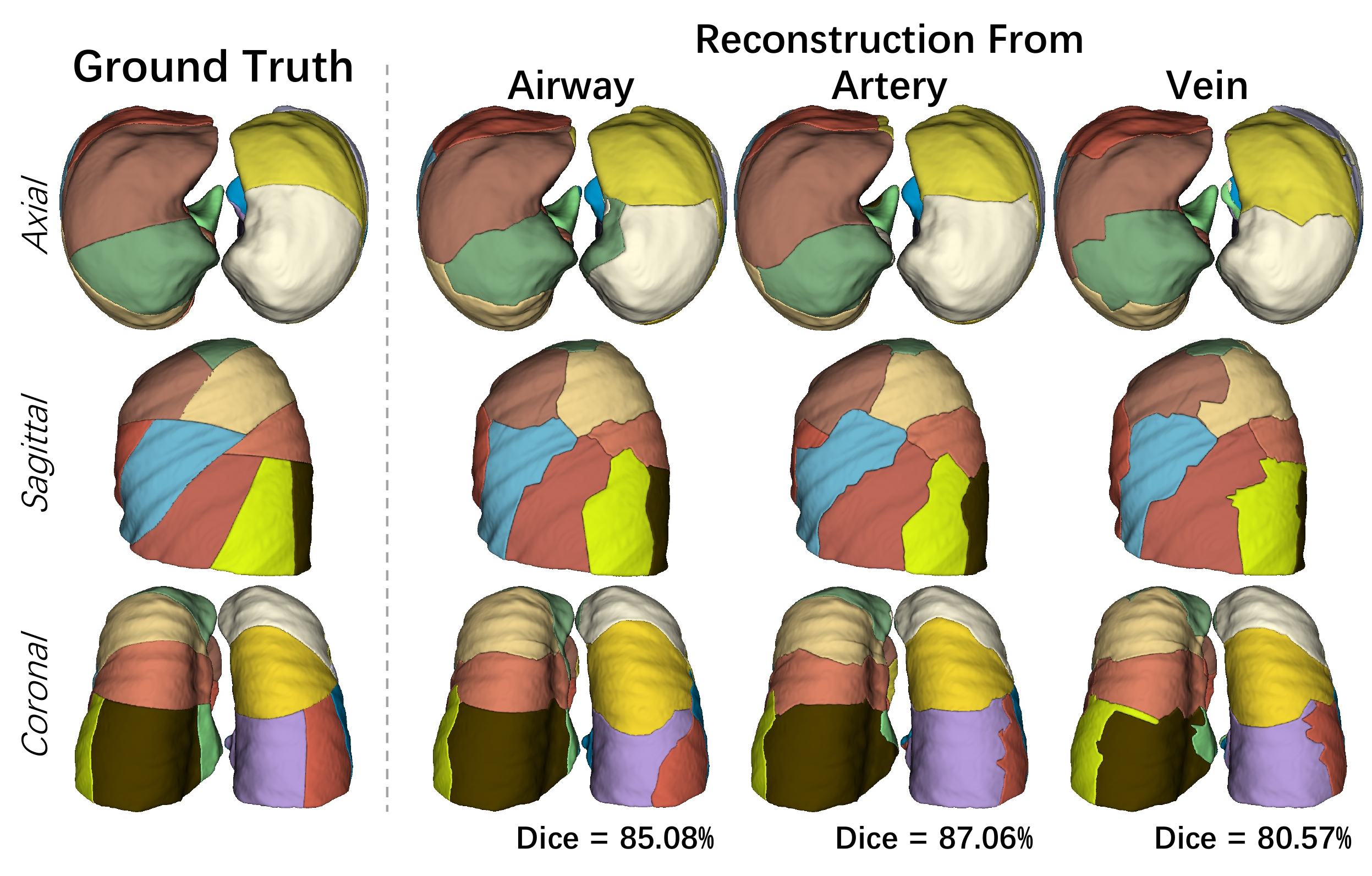}
	\caption{\textbf{Pulmonary Segments Visualization.} A sample reconstruction of pulmonary segments based on the airway, artery, and vein is presented against ground truth in axial, sagittal and coronal views.}
\label{fig:pulseg}
\end{figure}

\revised{
Pulmonary segments are subdivisions of the pulmonary lobes and are crucial for guiding segmentectomy, a widely applied surgical treatment for lung cancer with a higher survival rate than lobectomy~\citep{saji2022segmentectomy}. The definition of pulmonary segments is anatomically related to the pulmonary trees, rather than individual pixel texture features. On CT scans, while pulmonary lobes have visible boundaries, pulmonary segments do not. As illustrated in Fig.~\ref{fig:problem_formulation}, the spatial location of pulmonary trees implicitly defines the boundaries of pulmonary segments. A pulmonary segment must spatially contain the pulmonary trees of the same class, with boundaries between neighboring segments established along the inter-segmental veins~\citep{Kuang2022WhatMF}. Therefore, the segmentation of pulmonary trees is an important prerequisite for reconstructing pulmonary segments, not the other way around.

In the proposed architecture, the Implicit Point Module plays a vital role in defining implicit boundaries between different classes within the pulmonary tree, allowing for efficient dense reconstruction of the pulmonary tree. As the module learns a point-graph feature field for labeling, the input coordinates are not constrained to the tubular voxel/point in our target structures. Consequently, any point within the pulmonary lobe can be assigned a class label. By utilizing the extracted point-graph feature field from the airway, artery, or vein tree, our module can accurately infer class information for all points within the lobe, following a similar procedure described in Sec.~\ref{subsec:Imp_mod}.
}

Precisely, for each subject, we first acquire the point-based feature field from the proposed Point-Graph Network. As the binary mask for pulmonary lobes~(Fig.~\ref{fig:dataset} (a)) is given by the PTL dataset, we sample all foreground voxel/points in the mask, and iteratively perform query inferences on the Implicit Point Module based on the previously extracted feature field. This enables us to achieve a natural semantic reconstruction of the pulmonary segments, depicted in Fig. \ref{fig:pulseg}.

Compared to the ground truth, the tree-segmentation-induced pulmonary segments reconstruction achieves 79.89\%, 81.85\% and 76.16\% as micro-average dice scores, respectively for pulmonary airway, artery, and vein. Future works can potentially produce better pulmonary segment reconstruction by integrating features from the three pulmonary trees or leveraging the explicit pulmonary lobe boundaries as additional guidelines.

\section{Conclusion}
In conclusion, we take an experimentally comprehensive deep-dive into pulmonary tree segmentation based on the compiled PTL dataset. A novel architecture {Implicit Point-Graph Network (IPGN)} is presented for accurate and efficient pulmonary tree segmentation. Our method leverages a dual-branch point-graph fusion model to effectively capture the complex branching structure of the respiratory system. Extensive experiment results demonstrate that by implicit modeling on point-graph features, the proposed model achieves state-of-the-art segmentation quality with minimum computation cost for practical dense volume reconstruction. The advancements made in this study could potentially enhance the diagnosis, management, and treatment of pulmonary diseases, ultimately improving patient outcomes in this critical area of healthcare.

\section*{Data Availability}
The Pulmonary Tree Labeling (PTL) dataset and associated code can be accessed at \url{https://github.com/M3DV/pulmonary-tree-labeling}.

\section*{Acknowledgments}
This work was supported in part by the Swiss National Science Foundation.


\end{document}